\documentclass[conference]{template/IEEEtran}
\IEEEoverridecommandlockouts
\usepackage{cite}
\usepackage{mathtools}
\usepackage{amssymb,amsfonts}
\usepackage[linesnumbered,ruled]{algorithm2e}
\SetKwComment{Comment}{$\triangleright$\ }{}
\usepackage{graphicx}
\usepackage{textcomp}
\usepackage{xcolor}

\usepackage[normalem]{ulem}
\newif\ifshowtheorychanges
\showtheorychangesfalse
\newcommand{\theoryadd}[1]{%
  \ifshowtheorychanges\textcolor{blue}{#1}\else#1\fi}
\newcommand{\theorydel}[1]{%
  \ifshowtheorychanges
    \bgroup\markoverwith{\textcolor{red}{\rule[.5ex]{2pt}{.5pt}}}%
    \ULon{#1}%
  \fi}
\newsavebox{\theorystrikebox}
\newcommand{\theorydelmath}[1]{%
  \ifshowtheorychanges\mathpalette\theorystrike{#1}\fi}
\newcommand{\theorystrike}[2]{%
  \begingroup
  \sbox{\theorystrikebox}{$#1#2$}%
  \rlap{\raisebox{\dimexpr(\ht\theorystrikebox-\dp\theorystrikebox)/2\relax}{%
    \textcolor{red}{\rule{\wd\theorystrikebox}{.5pt}}}}%
  \usebox{\theorystrikebox}%
  \endgroup}

\newenvironment{theoryaddition}
  {\begingroup\ifshowtheorychanges\color{blue}\fi}
  {\par\endgroup}

\usepackage[utf8]{inputenc} % allow utf-8 input
\usepackage{url}            % simple URL typesetting
\usepackage{booktabs}       % professional-quality tables
\usepackage{amsthm}
\usepackage{subcaption}
\usepackage{bm}
\usepackage{xspace}
\usepackage[inline]{enumitem}
\usepackage{multirow}
\usepackage{multicol}
\usepackage{balance}

\def\BibTeX{{\rm B\kern-.05em{\sc i\kern-.025em b}\kern-.08em
    T\kern-.1667em\lower.7ex\hbox{E}\kern-.125emX}}

\usepackage{xparse}
\usepackage{etoolbox}
\let\en=\ensuremath

\newcommand{\dd}[1]{\mathop{}\!\mathrm{d} #1}

\DeclareMathOperator{\Ev}{\mathbb{E}}          % expectation
\DeclareMathOperator{\Cov}{\mathrm{Cov}}      % Covariance

\DeclarePairedDelimiter{\norm}{\lVert}{\rVert}
\DeclarePairedDelimiter{\abs}{\lvert}{\rvert}
\DeclarePairedDelimiter{\ip}{\langle}{\rangle}
\DeclareMathOperator{\Trace}{tr}

\DeclareMathOperator{\Rank}{rank}      % rank
\DeclareMathOperator{\diag}{diag}      % diag
\DeclareMathOperator{\blkdiag}{blockdiag}      % diag
\DeclareMathOperator{\vect}{vec}                % vectorization operator

\renewcommand{\vec}[1]{\en{\bm{\mathrm{#1}}}}

\newcommand{\hvec}[1]{\en{\hat{\vec{#1}}}}      % vector hat
\newcommand{\mat}[1]{\en{{\bm{\mathrm{#1}}}}}
\newcommand{\hmat}[1]{\en{\widehat{\mat{#1}}}}
\newcommand{\bmat}[1]{\en{\overline{\mat{#1}}}}
\newcommand{\tmat}[1]{\en{\widetilde{\mat{#1}}}}
\NewDocumentCommand{\grad}{O{}}{\nabla\ifblank{#1}{}{^{#1}}}
\newcommand{\tr}[0]{\en{\Trace}}

\newcommand{\R}[0]{\mathbb{R}}

\newcommand{\pdv}[2]{\frac{\partial #1}{\partial #2}}

\NewDocumentCommand{\E}{o m}{
  \Ev                          
  \IfValueT{#1}{_{#1}}
  \!\left[#2\right]
}
\newcommand{\Norm}[1]{\en{\mathcal{N}\left(#1\right)}}

\newtheorem{theorem}{Theorem}[section]

\newtheorem{lemma}[theorem]{Lemma}

\usepackage{pifont}% http://ctan.org/pkg/pifont

\begin{document}

\title{PAMS: Preconditioned Activation-Metric Subspace for
  Zeroth-Order LLM Fine-tuning}

% !!!
% Pradeep and Saksham, Fill in your information
\author{\IEEEauthorblockN{Pradeep Kumar Ragu Chanthar}
\IEEEauthorblockA{\textit{School of Computing} \\
\textit{University of Georgia}\\
Athens, USA \\
pradeep@uga.edu}
\and
\IEEEauthorblockN{Saksham Sahai Srivastava}
\IEEEauthorblockA{\textit{School of Computing} \\
\textit{University of Georgia}\\
Athens, USA \\
saksham.srivastava@uga.edu}
\and
\IEEEauthorblockN{Jaewoo Lee}
\IEEEauthorblockA{\textit{School of Computing} \\
\textit{University of Georgia}\\
Athens, USA \\
jaewoo.lee@uga.edu}
}

\maketitle

\begin{abstract}
Zeroth-order (ZO) optimization provides a computationally efficient approach to
fine-tuning large language models (LLMs), but conventional methods
typically rely on isotropic perturbations that ignore the anisotropic
structure of neural-network parameter spaces. We introduce
Preconditioned Activation-Metric Subspace (PAMS), a ZO
optimization framework that uses forward-pass activation statistics to
construct structured, layer-wise perturbation geometries without
computing gradients or forming dense curvature matrices. PAMS tracks a
low-rank approximation of the dominant activation subspace and its
associated spectrum online, and uses the spectrum to softly reweight
perturbation variance while preserving exploration in directions
outside the retained subspace. A trace normalization separates the
perturbation geometry from its overall scale, enabling consistent
perturbation magnitudes across layers. We show theoretically that the
resulting perturbation covariance induces right-side preconditioning
of the gradient of the corresponding Gaussian-smoothed objective. We
further characterize the estimator's smoothing bias and second moment
and derive a nonconvex convergence guarantee that captures the effects
of the perturbation geometry and smoothing radius. The proposed
low-rank tracking procedure requires $\mathcal{O}(n_lr)$ persistent
memory for a layer with input dimension $n_l$ and retained rank $r$,
avoiding the $\mathcal{O}(n_l^2)$ memory required by a dense activation
covariance. Experiments on transformer-based language-model
fine-tuning tasks across GLUE and SuperGLUE demonstrate that PAMS
improves the effectiveness and stability of zeroth-order optimization
while retaining its gradient-free and objective-evaluation-based
advantages.

%%% Local Variables:
%%% mode: latex
%%% TeX-master: "../main"
%%% End:

\end{abstract}

\begin{IEEEkeywords}
Zeroth-Order Optimization, Subspace Optimization, Large Language Models
\end{IEEEkeywords}

\section{Introduction}
\label{sec:introduction}
% Large language models (LLMs) have emerged as the foundational paradigm
% of modern artificial intelligence, demonstrating remarkable
% capabilities across a wide range of tasks including natural language
% understanding, reasoning, code generation, and multimodal
% learning.
Fine-tuning large language models (LLMs) with first-order methods such
as stochastic gradient descent 
(SGD)~\cite{robbins1951Stochastic} and
AdamW~\cite{Loshchilov2019DecoupledWD} requires
backpropagation and, in typical implementations, storage of
intermediate activations can impose substantial memory and
computational overhead~\cite{rajbhandari2020zero}.
Zeroth-order (ZO) optimization offers an alternative by estimating
update directions from objective evaluations, without explicitly
computing parameter gradients. This makes ZO methods
attractive for memory-constrained fine-tuning and
settings in which gradient access is unavailable. 
A central limitation of conventional ZO methods is the geometry of
perturbation sampling.
Conventional ZO methods typically draw perturbations from an isotropic
distribution, assigning equal variance to all parameter
directions. Such a geometry ignores the strong anisotropy present in
neural-network parameter spaces, where different feature directions
can have substantially different scales and influence the model
computation differently~\cite{liu2020understanding,park2022how,yu2025zeroth}.
As a result, a fixed isotropic perturbation distribution may allocate
a substantial fraction of its perturbation budget to directions that
provide relatively little useful information for optimization. The
resulting estimator can be statistically
inefficient~\cite{Ghadimi2013StochasticFA}, increasing the 
number of objective evaluations required to obtain a useful
update. The challenge is therefore to introduce informative anisotropy
into ZO exploration without relying on the gradients, Hessian, or
Fisher information that gradient-free optimization seeks to avoid. 
% Second, modern transformer architectures
% exhibit highly anisotropic optimization landscapes with
% ill-conditioned curvature. This pathological geometry induce severe training
% instability, sensitivity to hyperparameters, and inefficient
% exploration of parameter
% space~\cite{liu2020understanding,park2022how}.
We observe that useful structural information is available directly
from forward computation. Each forward pass produces intermediate
activations whose second-moment statistics characterize the feature
directions represented by the current data and model.
These statistics exhibit substantial structure and anisotropy in
neural networks~\cite{gao2019representation,godey2024anisotropy}, 
suggesting that they can provide a useful source of geometry for ZO
exploration. This motivates the question: can forward-pass activation
statistics define an informative perturbation geometry
while preserving the gradient-free nature of zeroth-order optimization? 

We answer this question with Preconditioned Activation-Metric Subspace (PAMS), a
zeroth-order optimization framework that uses forward-pass activation
statistics to construct spectrum-aware perturbations. For each layer,
PAMS identifies a low-rank approximation to the dominant activation
subspace and uses the corresponding activation spectrum to reweight
perturbation variance. Directions associated with 
larger activation second moments receive smaller perturbations, while
directions outside the retained subspace remain isotropic. Thus, PAMS
does not restrict exploration to a fixed low-dimensional subspace;
instead, it uses the activation spectrum to redistribute the
perturbation budget while preserving exploration in the complementary
directions. A trace normalization keeps the total perturbation
variance fixed, separating the choice of exploration geometry from the
overall perturbation scale.
This construction also admits a direct optimization
interpretation. For a general ZO estimator, the covariance
of the perturbation distribution determines the preconditioning
applied to the gradient of the corresponding Gaussian-smoothed
objective~\cite{Nesterov2017RandomGM}.
PAMS therefore uses forward-pass activation statistics not merely as a
heuristic for selecting promising perturbation directions, but as a
structured covariance operator that determines the geometry of the
expected ZO update. In PAMS, this covariance is layer-specific and
anisotropic along the input-feature dimension, yielding right-side
preconditioning of the gradient of the globally smoothed
objective. This provides a direct connection between information
available from forward computation and the optimization geometry of
ZO estimation. 
To make this construction practical for large transformer layers, PAMS
maintains the activation geometry online without explicitly
forming activation second-moment matrices. It tracks the dominant
activation subspace from mini-batch activations and estimates the
associated spectrum within the retained subspace.
For a layer with input dimension $n_l$ and retained rank $r$, the
persistent state requires $\mathcal{O}(n_lr)$ memory rather than
$\mathcal{O}(n_l^{2})$, making the procedure practical when $r\ll
n_l$. The resulting low-rank representation can also be applied
directly when sampling perturbations, avoiding the construction of
dense parameter-space covariance matrices.
We provide a theoretical analysis of PAMS from both the optimization
and estimator perspectives. We show that the structured perturbation
covariance induces right-side preconditioning of the gradient of the
Gaussian-smoothed objective. We further characterize the smoothing
bias and the second moment of the resulting zeroth-order estimator,
yielding a nonconvex convergence guarantee that explicitly captures
the effects of the perturbation geometry and smoothing radius.
These results clarify both the optimization benefit and the
statistical cost of using structured, activation-informed
perturbations.
% !!! Commented out as this is reiterated in Section V
%
% Detailed proofs of our theoretical results and additional
% technical details are provided in the appendix of the
% arXiv version of this paper.
%
We evaluate PAMS on transformer-based language-model fine-tuning
tasks, including GLUE~\cite{Wang2018GLUEAM} and
SuperGLUE~\cite{Wang2019SuperGLUEAS} benchmarks. Our experiments 
demonstrate that incorporating activation-informed geometry can
substantially improve the stability and query efficiency of
ZO optimization while preserving its memory and evaluation
advantages.

Our contributions are summarized as follows:
\begin{enumerate}[leftmargin=*]
\item \textbf{Activation-informed perturbation geometry.} We introduce PAMS,
  which uses forward-pass activation statistics to construct an
  anisotropic perturbation distribution for ZO optimization. 
\item \textbf{Spectrum-aware soft preconditioning.} PAMS uses the activation
  spectrum to continuously reweight exploration while retaining
  exploration outside the dominant activation subspace, providing a
  flexible alternative to isotropic perturbations. 
\item \textbf{Efficient online subspace tracking.} We develop a low-rank
  procedure for tracking activation geometry and its spectrum without
  explicitly forming large activation second-moment matrices.   
\item \textbf{Zeroth-order preconditioning and convergence analysis.} We establish the
  connection between the perturbation covariance and preconditioning
  of the Gaussian-smoothed objective, together with convergence
  guarantees that characterize the resulting smoothing and estimation
  errors. 
\item \textbf{Empirical evaluation for LLM fine-tuning.}
  We evaluate PAMS across GLUE and SuperGLUE fine-tuning tasks and
  show that it achieves competitive or superior performance relative
  to existing ZO optimizers while preserving the practical advantages
  of gradient-free optimization. 
\end{enumerate}

%%% Local Variables:
%%% mode: latex
%%% TeX-master: "../main"
%%% End:

\section{Related Work}
\label{sec:related}
Zeroth-order (ZO) optimization has emerged as a memory-efficient alternative 
to backpropagation for adapting LLMs. Malladi et al.~\cite{malladi2023finetuning}
introduced \textsc{MeZO}, an in-place ZO-SGD method whose memory
footprint is comparable to inference, establishing the feasibility of
fine-tuning large language models using only forward passes. However,
its isotropic full-dimensional perturbations do not account for the
highly nonuniform geometry of the loss landscape and can suffer from
high estimator variance in large parameter spaces. To better exploit
this geometry, Zhao et al.~\cite{zhao2025secondorder} proposed
\textsc{HiZOO}, which incorporates diagonal Hessian information to
adaptively rescale ZO updates and account for heterogeneous curvature
across parameter dimensions. Seung et al.~\cite{seung2026low} further
introduced \textsc{LOREN}, which captures richer curvature structure
through a memory-efficient low-rank block-diagonal preconditioner,
reformulates ZO preconditioning as learning an anisotropic
perturbation distribution, and employs a variance-reduced gradient
estimator. Collectively, these works demonstrate the importance of
moving beyond isotropic random perturbations by incorporating
informative structure into the design and scaling of ZO search
directions, motivating more structured and data-informed approaches to
ZO fine-tuning.   

A complementary direction addresses the unfavorable dimensional
dependence of ZO gradient estimation by restricting perturbations to
structured low-dimensional spaces. Chen et
al.~\cite{chen2025enhancing} proposed \textsc{LOZO}, which exploits
the low-rank structure observed in LLM fine-tuning gradients and
performs optimization over randomly sampled low-dimensional
subspaces. Similarly, Yu et al.~\cite{yu2025zeroth} introduced
\textsc{SubZero}, which uses layer-wise low-rank random subspaces to
reduce the effective perturbation dimension and variance of ZO
gradient estimates. These works show that the structure of the
perturbation space is critical for scalable ZO fine-tuning. However,
their low-dimensional subspaces are primarily determined through
random constructions rather than data-dependent information available
during the forward pass. In contrast, PAMS tracks the dominant
low-rank activation subspace from forward-pass statistics and uses its
spectrum to anisotropically shape the perturbation distribution, while
preserving isotropic exploration in the complementary directions.  

Another direction considers heterogeneity across layers or uses
activation statistics to construct efficient preconditioners.
Tan et al.~\cite{tan2026harmony} analyzed the layer-wise divergence between
first- and zeroth-order optimization and proposed \textsc{DiZO}, which
adapts update magnitudes across layers through projection-based
scaling. DiZO demonstrates that accounting for layer-wise differences
can improve the efficiency of ZO fine-tuning. 
Relatedly, Seung et al.~\cite{seung2025mac} proposed
\textsc{MAC}, which uses activation statistics to construct efficient
curvature approximations for preconditioning. Although \textsc{MAC}
operates in a first-order setting and also relies on gradient-related
quantities, it demonstrates that activation-derived factors can be
effectively exploited in the design of computationally efficient
preconditioners. PAMS builds on these complementary insights by
combining low-rank activation structure with activation-informed
preconditioning. It tracks the dominant eigenspace of the activation
second-moment matrix and uses its spectrum to construct a structured
perturbation covariance for ZO optimization. Thus, PAMS uses
activation-derived structure to identify the directions in which ZO
perturbations should be spectrally reweighted, while retaining
isotropic exploration in the complementary subspace and the memory
advantages of forward-only fine-tuning.

%%% Local Variables:
%%% mode: latex
%%% TeX-master: "../main"
%%% End:

\section{Preliminaries}
\label{sec:prelim}
%SPSA~\cite{Spall1992MultivariateSA}

\subsection{Problem Setup and Notation}
We consider the optimization of a parameterized model with parameters
$\mat{W}=\{\mat{W}_l\}_{l=1}^{L}$ over the stochastic objective  
\begin{equation}
    \label{eq:prelim_objective}
    \mathcal{L}(\mat{W})
    = \mathbb{E}_{\xi}\!\left[\ell(\mat{W};\xi)\right],
\end{equation}
where $\xi$ denotes a data sample or mini-batch and $\ell(\cdot)$ is a
loss function. For a linear layer
$l$, let $\mat{W}_l\in\R^{m_l\times n_l}$ denote its weight matrix,
where $n_l$ and $m_l$ are the input and output dimensions,
respectively.
We use $\vec{w}=\vect(\mat{W}_1,\ldots,\mat{W}_L) \in \R^{d}$ to
denote the vectorization of all model parameters and $d =
\sum_{l=1}^{L}m_ln_l$.
For a matrix
$\mat{X}$, $\vect(\mat{X})$ denotes column-wise vectorization,
$\mat{X}^{\intercal}$ its transpose, 
$\tr(\mat{X})$ its trace, and $\mat{I}_k$ the $k\times k$ identity
matrix. We use $\otimes$ to denote the Kronecker product. The
symmetric part of a square matrix $\mat{X}$ is denoted by
$\mathrm{sym}(\mat{X})=\frac{1}{2}(\mat{X}+\mat{X}^{\intercal})$.

\subsection{Matrix-Normal Distribution}
The matrix-normal distribution is a convenient way to represent
Gaussian random matrices whose covariance separates across their two
matrix dimensions. For a matrix $\mat{\Delta}\in \R^{m\times n}$, we write
$\mat{\Delta}\sim \mathcal{MN}(\mat{M}, \mat{R}, 
\mat{C})$, where $\mat{\Delta}$ has mean $\mat{M}$, row covariance
$\mat{R}\in \R^{m\times m}$, and column covariance $\mat{C}\in
\R^{n\times n}$.
The key property of the matrix-normal distribution is its equivalence
to a multivariate Gaussian after vectorization~\cite{gupta1999matrix}:
\[
  \mat{\Delta}\sim\mathcal{MN}(\mat{M}, \mat{R}, \mat{C})
  \Leftrightarrow \vect(\mat{\Delta}) \sim \Norm{\vect(\mat{M}), \mat{C}\otimes \mat{R}}\,.
\]
Thus, the Kronecker product $\mat{C}\otimes \mat{R}$ specifies the
covariance of the vectorized matrix while preserving its separable
row-column structure. A matrix-normal random variable can be sampled
as
\[
  \mat{\Delta} = \mat{M} + \mat{R}^{1/2}\mat{Z}\mat{C}^{1/2}\,,
\]
where the entries of $\mat{Z}\in \R^{m\times n}$ are independent standard Gaussian
variables, i.e., $Z_{ij} \overset{\mathrm{i.i.d.}}{\sim}\Norm{0, 1}$.
This representation is useful when different covariance structures are
desired along the two dimensions of a matrix. In particular, it allows
structured Gaussian perturbations to be represented through a
matrix-valued random variable without explicitly constructing the full
covariance matrix of its vectorized form. 

\subsection{Stiefel Manifold and Riemannian Optimization}
The Stiefel manifold is the set of matrices with orthonormal
columns. For $\mat{U}\in \R^{n\times r}$, it is defined as
\[
  \mathrm{St}(r, n) = \{\mat{U}\in \R^{n\times
    r}~:~\mat{U}^{\intercal}\mat{U} = \mat{I}_r\}\,.
\]
An element $\mat{U}$ therefore represents an $r$-dimensional subspace
together with a particular orthonormal basis.
When optimizing a differentiable function $f(\mat{U})$ over the
Stiefel manifold, the Euclidean gradient must be projected onto the
tangent space at $\mat{U}$. Under the standard embedded Euclidean metric,
this projection is~\cite{absil2008optimization}
\[
  \mathrm{grad}f(\mat{U}) = \grad{f}(\mat{U}) -
  \mat{U}\mathrm{sym}(\mat{U}^{\intercal}\grad{f}(\mat{U}))\,.
\]
A Riemannian gradient ascent step with step size $\eta_U>0$ is then
\[
  \tmat{U} = \mat{U} + \eta_U\mathrm{grad}f(\mat{U})\,.
\]
Because $\tmat{U}$ does not generally have orthonormal columns, it is
mapped back to the Stiefel manifold using a retraction. We use the QR
retraction, $\mat{U}^{+} = \mathrm{qf}(\tmat{U})$,
where $\mathrm{qf}(\cdot)$ denotes the orthonormal $Q$ factor of a QR decomposition.
This gives the standard Riemannian gradient-update procedure used to
maintain an orthonormal basis. In
Section~\ref{sec:subspace_tracking}, we apply this procedure to the
PCA objective defined on the Stiefel manifold to 
track the dominant activation subspace online.

\section{Preconditioned Activation-Metric Subspace}
\label{sec:method}
% In this section, we formalize the architecture of the proposed Preconditioned
% Activation-Metric Subspace (PAMS) framework.
% PAMS performs zeroth-order optimization using a learned
% activation-induced metric that can be expressed as a low-rank
% deformation of the Euclidean geometry. The deformation is estimated
% online from activation statistics and concentrates geometric
% adaptation within a small activation subspace while preserving
% isotropic exploration in all remaining directions.
We introduce Preconditioned Activation-Metric Subspace (PAMS), a
zeroth-order optimization framework that uses forward-pass activation
statistics to define layer-wise perturbation geometries. For each
layer, PAMS constructs a low-rank activation-informed preconditioner,
normalizes its scale, and samples zeroth-order perturbations from the
resulting structured covariance. The required activation
subspaces and their associated spectra are maintained online from
mini-batch activations, without explicitly forming the activation
second-moment matrices. The resulting procedure is described in
Algorithm~\ref{alg:pams}. 

\subsection{Activation-Informed Perturbation Geometry}
\label{sec:metric_construction}
% A central challenge in zeroth-order optimization is the construction
% of perturbation distributions that efficiently explore
% high-dimensional parameter spaces. Standard zeroth-order methods
% employ isotropic Gaussian perturbations, implicitly assuming that all
% parameter directions are equally informative. In large transformer
% models, however, both activations and gradients exhibit highly
% anisotropic structure, causing isotropic exploration to allocate
% excessive variance to directions with little influence on the
% objective.
%
% To motivate a geometry-aware perturbation strategy,
PAMS constructs a perturbation geometry from statistics that are
observable during the forward pass. 
Consider a linear layer $l$
with weight matrix $\mat{W}_l \in \R^{m_l \times n_l}$,
input activation matrix $\mat{A}_l \in \R^{n_l \times b}$, and
pre-activation output $\mat{Z}_l = \mat{W}_l\mat{A}_l\in \R^{m_l\times b}$,
where $b$ denotes the mini-batch
size and $n_l$ is the input feature dimension.
Let $\mathcal{L}$ denote the training objective and define
$\mat{Y}_l = \pdv{\mathcal{L}}{\mat{Z}_l}$
as the corresponding pre-activation gradient. 
The layer gradient of $\mathcal{L}$ with respect 
to $\mat{W}_l$ is $\grad_{\mat{W}_l} \mathcal{L} =
\mat{Y}_l\mat{A}_l^{\intercal}$.
Under the standard Kronecker-factored
approximation used by K-FAC~\cite{martens2015optimizing}, the layer-wise Fisher
matrix can be approximated as $\mat{F}_l\approx \mat{\Sigma}_A \otimes \mat{\Sigma}_Y$,
where $\mat{\Sigma}_A = \E{\mat{A}\mat{A}^{\intercal}}$ and
$\mat{\Sigma}_Y = \E{\mat{Y}\mat{Y}^{\intercal}}$ denote the
second-moment factors of the activations and the pre-activation
gradients, respectively.
The two factors have fundamentally different information
requirements. The activation second-moment matrix $\mat{\Sigma}_{A,l}$ is
directly observable from forward activations, whereas
$\mat{\Sigma}_{Y,l}$ depends on backpropagated gradients and is
therefore unavailable in a gradient-free optimization setting.
PAMS therefore retains only the forward-observable factor and uses
an identity operator for the complementary output dimension: 
\begin{equation}
  \label{eq:pams_approx}
  \mat{M}_l= \mat{\Sigma}_A\otimes \mat{I}_{m_l}\,.
\end{equation}
Equation~\eqref{eq:pams_approx} is a deliberate factor-selection
design, rather than an assumption that the output-side Fisher
factor $\mat{\Sigma}_{Y,l}$ is
intrinsically isotropic.
PAMS deliberately captures the anisotropy that can be inferred from
forward activations while leaving the gradient-dependent component
isotropic. This construction provides a principled way to incorporate
activation geometry into ZO exploration without estimating
gradients, the full Fisher matrix, or a dense parameter-space
curvature matrix. 
This design choice is further motivated by empirical
evidence that
activation-side preconditioning can capture substantial optimization
benefits even when the complementary K-FAC factor is
omitted~\cite{Benzing2022GradientDO}. 

Let the eigendecomposition of the activation second-moment matrix be
$\mat{\Sigma}_{A,l} = \mat{U}_l\mat{\Lambda}_l\mat{U}_l^{\intercal}$
with eigenvalues $\lambda_{l,1}\geq \cdots \geq \lambda_{l,n_l}\geq 0$.
Let $\mat{U}_{l,r}$ contain the leading $r$ eigenvectors and let
$\mat{\Lambda}_{l,r} = \diag(\lambda_{l,1},
\ldots,\lambda_{l,r})$. PAMS defines the activation preconditioner
\begin{equation} \label{eq:ai_preconditioner}
  \mat{P}_l =
  \mat{U}_{l,r}(\mat{\Lambda}_{l,r}+\rho\mat{I}_r)^{-\alpha}\mat{U}_{l,r}^{\intercal}
  + (\mat{I} -\mat{U}_{l,r}\mat{U}_{l,r}^{\intercal})\,,
\end{equation}
where $\rho > 0$ is a damping parameter and $\alpha \in [0, 1]$
controls the strength of preconditioning. Equivalently, 
\begin{equation}
  \label{eq:pl_deformation}
    \mat{P}_l = \mat{I} +
  \mat{U}_{l,r}\left[(\mat{\Lambda}_{l,r}+\rho\mat{I}_r)^{-\alpha}-\mat{I}\right]
  \mat{U}_{l,r}^{\intercal}\,.
\end{equation}
This construction has two desirable properties. First,
$\Rank(\mat{P}_l - \mat{I}) \leq r$, so PAMS modifies the geometry
only within the retained activation subspace. Second, along a retained
eigenvector $\vec{u}_{l,i}$, the corresponding eigenvalue of
$\mat{P}_l$ is $(\lambda_{l,i} +\rho)^{-\alpha}$, whereas
every direction orthogonal to the retained subspace has eigenvalue
$1$. Hence directions with larger activation second moments receive
stronger relative attenuation, while directions outside the retained
subspace remain isotropic. The damping parameter $\rho$ prevents
excessive spectral reweighting when an activation eigenvalue is small.
The parameter $\alpha$ interpolates between isotropic and
activation-informed exploration. When $\alpha=0$, $\mat{P}_l =
\mat{I}_{n_l}$, and the perturbation geometry is isotropic.
Increasing $\alpha$ increases the dependence of the perturbation
geometry on the activation spectrum, with the limiting case $\alpha=1$
corresponding to a damped inverse-spectrum weighting within the
retained subspace. Importantly, PAMS does not claim that $\mat{P}_l$
recovers the full Fisher or Hessian geometry. Instead, it extracts the
portion of the layer-wise geometry that is accessible from
forward-pass activation statistics and represents that information
through a low-rank deformation of the identity.
The matrix $\mat{P}_l$ therefore specifies the activation-informed
geometry for ZO exploration.
In the next subsection, we convert this geometry into a perturbation
distribution while controlling its overall scale and avoiding the
construction of dense $n_l\times n_l$ matrices. 

\subsection{Geometry-Aware Perturbation Sampling}
\label{sec:perturbation_sampling}
The activation preconditioner $\mat{P}_l$ determines how zeroth-order
exploration is distributed across the input-feature directions of
layer $l$.
Because the scale of the activation spectrum can vary substantially
across layers and throughout training, directly using $\mat{P}_l$
would result in layer-dependent perturbation magnitudes. We therefore
normalize each preconditioner to preserve a common perturbation scale. 
\begin{equation} \label{eq:normalization}
  \bmat{P}_l = \frac{n_l}{\tr(\mat{P}_l)}\mat{P}_l\,.
\end{equation}
By construction, $\tr(\bmat{P}_l) = n_l$, so the average marginal
variance across the $n_l$ input-feature directions is identical to
that of an $n_l$-dimensional standard isotropic Gaussian.
Importantly, this normalization changes only the overall
scale of $\mat{P}_l$ and therefore preserves its eigenvectors and
relative spectral structure.
Because of the low-rank form in~\eqref{eq:pl_deformation}, its trace is
available directly from the retained activation spectrum:
\[
  \tr(\mat{P}_l) = (n_l - r) + \sum_{i=1}^{r}(\lambda_{l,i}+\rho)^{-\alpha}\,.
\]
Thus, the normalization requires only $\mathcal{O}(r)$
computation.
For each layer, we define the corresponding vectorized perturbation covariance as
\[
  \mat{C}_l = \bmat{P}_l\otimes \mat{I}_{m_l}\,.
\]
PAMS samples perturbations for all layers
simultaneously. Specifically,
\[
  \mat{\Delta}_l \sim\mathcal{MN}(\mat{0}, \mat{I}_{m_l},
  \bmat{P}_l)\,, \quad l=1, \ldots, L\,,
\]
independently across layers.
Equivalently, the vectorized perturbation has covariance
$\Cov[\vect(\mat{\Delta}_l)] = \bmat{P}_l\otimes \mat{I}_{m_l}$.
Within each layer, the covariance is anisotropic along the
input-feature dimension according to the activation geometry, while
remaining isotropic along the output dimension. The independence
across layers allows each $\mat{\Delta}_l$ to be sampled and applied
independently without constructing a dense covariance over the full
parameter space.
The perturbation can be generated efficiently using the low-rank
representation of each $\bmat{P}_l^{1/2}$. In particular, for 
$\mat{R}_l\in \R^{m_l\times n_l}$ whose entries $R_{l,
  ij}\sim\Norm{0,1}$ are 
independent Gaussians, PAMS generates $\mat{\Delta}_l =
\mat{R}_l\bmat{P}_l^{1/2}$.
From the construction in Section~\ref{sec:metric_construction}, 
the square root of $\bmat{P}_l$ has the closed-form expression
\[
  \bmat{P}_l^{1/2} = \sqrt{\frac{n_l}{\tr(\mat{P}_l)}}\left[
    \mat{I} + \mat{U}_{l,r} \left((\mat{\Lambda}_{l,r} + \rho
      \mat{I}_r)^{-\frac{\alpha}{2}}-\mat{I}_r\right)
    \mat{U}_{l,r}^{\intercal} \right]\,.
\]
Therefore, multiplication by $\bmat{P}_l^{1/2}$ can be implemented
using the retained $r$-dimensional subspace without forming an $n_l\times n_l$
dense matrix.

\paragraph{Antithetic ZO Estimation}
% \theoryadd{We write $\ell(\mat{W};\xi)$ for the mini-batch loss, with
% $\mathcal{L}(\mat{W})=\E[\xi]{\ell(\mat{W};\xi)}$.
% Both antithetic evaluations share the same mini-batch $\xi$, including
% any randomness in the loss. The covariance is fixed before drawing
% fresh Gaussian perturbations.}
Given a sampled perturbation $\{\mat{\Delta}_l\}_{l=1}^{L}$, PAMS constructs
two perturbed networks
\[
  \mat{W}^{+} = \{\mat{W}_l + \varepsilon\mat{\Delta}_l\}_{l=1}^{L}\,,\quad
  \mat{W}^{-} = \{\mat{W}_l - \varepsilon\mat{\Delta}_l\}_{l=1}^{L}\,.
\]
and evaluates the mini-batch loss once at each perturbed network state. The same
pair of function evaluations provides a ZO gradient estimate for every
layer: 
\[
  \hmat{G}_l = \frac{\ell(\mat{W}^{+};\xi) -
        \ell(\mat{W}^{-};\xi)}{2\varepsilon}\mat{\Delta}_l\,,\quad l
  =1, \ldots, L\,.
\]
Thus, a single antithetic pair of full-network evaluations yields the
perturbation estimates for all layers simultaneously, irrespective of
the number of layers.
To characterize the effect of the perturbation covariance, let 
$\vec{w} = \vect(\mat{W}_1, \ldots, \mat{W}_L)$ and $\vec{u} =
\vect(\mat{\Delta}_1, \ldots, \mat{\Delta}_L)$ so that 
$\vec{u}\sim\Norm{\vec{0}, \mat{C}}$, where $\mat{C} =
\blkdiag(\mat{C}_1, \ldots, \mat{C}_L)$.
For a smoothing radius $\varepsilon>0$, define the Gaussian smoothed
objective $\mathcal{L}_{\varepsilon, \mat{C}} = \E[\vec{u}\sim\Norm{\vec{0}, \mat{C}}]{\mathcal{L}(\vec{w}
  + \varepsilon\vec{u})}$.
The following result establishes the connection between the
perturbation covariance and ZO preconditioning. 
\begin{lemma}[Perturbation covariance induces zeroth-order
  preconditioning]
  \label{lem:zo_precond}
  Let $\mathcal{L}~:~\R^{d}\to \R$ be differentiable %\theoryadd{with a globally Lipschitz gradient}
  , and let
  $\vec{u}\sim\Norm{\vec{0}, \mat{C}}$, where $\mat{C}\succeq \mat{0}$. For a smoothing radius
  $\varepsilon>0$, define the Gaussian-smoothed objective
  $\mathcal{L}_{\varepsilon,\mat{C}}(\vec{w}) =
  \E[\vec{u}]{\mathcal{L}(\vec{w}+\varepsilon\vec{u})}$. Consider the
  two-sided antithetic ZO estimator
  \[
    \hvec{g}_{\varepsilon, \mat{C}}(\vec{w}; \vec{u}) =
    \frac{\mathcal{L}(\vec{w}+\varepsilon\vec{u}) -
      \mathcal{L}(\vec{w}-\varepsilon\vec{u})}{2\varepsilon} \vec{u}\,.
  \]
  Then $\E[\vec{u}]{\hvec{g}_{\varepsilon,\mat{C}}(\vec{w};\vec{u})} =
  \mat{C}\grad{\mathcal{L}}_{\varepsilon, \mat{C}}(\vec{w})$.
\end{lemma}
\begin{proof}
  Taking the expectation of the estimator gives
  \begin{align}
    \E[\vec{u}]{\hvec{g}_{\varepsilon,\mat{C}}(\vec{w};\vec{u})}
    &=\frac{1}{2\varepsilon}\E[\vec{u}]{\left(\mathcal{L}(\vec{w}+\varepsilon\vec{u}) -
      \mathcal{L}(\vec{w}-\varepsilon\vec{u})\right)\vec{u}}\nonumber \\
    &=\frac{1}{\varepsilon}
      \E[\vec{u}]{\mathcal{L}(\vec{w}+\varepsilon\vec{u})\vec{u}}\,, \label{eq:exp_pams1}
  \end{align}
  where the last equality uses the symmetry of the Gaussian
  distribution, 
  $\E[\vec{u}]{\mathcal{L}(\vec{w}-\varepsilon\vec{u})\vec{u}}=-\E[\vec{u}]{\mathcal{L}(\vec{w}+\varepsilon\vec{u})\vec{u}}$.
  For $\vec{u}\sim\Norm{\vec{0}, \mat{C}}$, Stein's identity states that, for
  any sufficiently regular scalar function $f$, $\E[\vec{u}]{\vec{u}f(\vec{u})} =
  \mat{C}\E[\vec{u}]{\grad_{\vec{u}}f(\vec{u})}$.
  Applying this identity to $f(\vec{u}) =
  \mathcal{L}(\vec{w}+\varepsilon\vec{u})$, we obtain
  \begin{align*}
    \E[\vec{u}]{\hvec{g}_{\varepsilon,\mat{C}}(\vec{w};\vec{u})}
    &=\frac{\mat{C}}{\varepsilon}\E[\vec{u}]{\grad_{\vec{u}}\mathcal{L}(\vec{w}+\varepsilon\vec{u})}\\
    &=\frac{\mat{C}}{\varepsilon}\E[\vec{u}]{\varepsilon\grad_{\vec{w}}\mathcal{L}(\vec{w}+\varepsilon\vec{u})}   \\
    &=\mat{C}\grad_{\vec{w}}\E[\vec{u}]{\mathcal{L}(\vec{w}+\varepsilon\vec{u})}\,,
  \end{align*}
  where the penultimate equality follows from the chain rule and the
  final equality follows by differentiating under
  expectation. Therefore,
  $\E[\vec{u}]{\hvec{g}_{\varepsilon,\mat{C}}(\vec{w};\vec{u})} =
  \mat{C}\grad_{\vec{w}}\mathcal{L}_{\varepsilon, \mat{C}}(\vec{w})$.
\end{proof}
Because $\mat{C}$ is block diagonal, the layer-wise component of the
resulting expectation is
\begin{equation} \label{eq:right_side_precond}
    \E{\hmat{G}_l} = \grad_{\mat{W}_l}\mathcal{L}_{\varepsilon,
      \mat{C}}\bmat{P}_l\,, \quad l=1, \ldots, L\,.
\end{equation}
Thus, although the perturbation is sampled jointly across all layers
and the objective is evaluated on the full perturbed network, each
layer inherits its own activation-informed right-side
preconditioning of the gradient of the globally smoothed objective.
\theoryadd{For mini-batch evaluations, this population identity holds
conditionally when the covariance is selected independently of the fresh
loss batch given the past. If the covariance uses that batch, the
conditional identity instead applies to its sample loss.
Section~\ref{sec:analysis} addresses the resulting dependence,
and further details are provided in Appendix~\ref{app:preconditioning}.
Equation~\eqref{eq:right_side_precond} establishes the central
mechanism of PAMS. The covariance of the ZO perturbation directly
determines the preconditioning applied to the gradient of the
corresponding smoothed objective. The trace normalization
controls the overall perturbation scale, while the relative eigenvalue
structure of the activation second-moment matrix determines how
exploration is distributed within each layer.
The use of a joint perturbation across all layers enables the same two
objective evaluations to provide ZO estimates for every parameter
block simultaneously. In combination with the antithetic construction,
PAMS therefore requires exactly two full-network objective evaluations
per optimization iteration, independent of the number of layers.
Importantly, while the estimator follows the
SPSA~\cite{Spall1992MultivariateSA} paradigm, PAMS differs in that the perturbation covariance is
structured and layer-specific, rather than isotropic, allowing all
layers to share the same pair of forward evaluations while still preserving
per-layer geometric preconditioning.
The additional computational overhead is limited to constructing the
perturbations and applying the low-rank activation subspaces, which
scales linearly with the retained rank.}

\begin{algorithm}[t]
  \caption{PAMS: Preconditioned Activation-Metric Subspace Optimization}
  \label{alg:pams}
  \DontPrintSemicolon

  \KwIn{
    Initial parameters $W_0$,
    rank $r$,
    finite-difference radius $\varepsilon$,
    learning rate $\eta$,
    subspace learning rate $\eta_U$,
    damping $\rho$,
    preconditioning exponent $\alpha$,
    exploration budget $\tau$,
    EMA coefficient $\beta_\Lambda$
  }
  Initialize orthonormal basis $\mat{U}_0 \in \R^{n_l \times r}$\;
  Initialize spectrum estimate $\mat{\Lambda}_0= \mat{I}_r$\;
  \For{$t=0,1,\ldots,T-1$}{
    \theorydel{Obtain mini-batch activations $A_t$}
    \theoryadd{Draw a fresh mini-batch $\xi_t$ and obtain its activations $\mat{A}_t$}\;
    $\mat{G}_t \gets \frac{1}{b}\mat{A}_t(\mat{A}_t^{\intercal}\mat{U}_t)$\;
    $\mathrm{grad}f(\mat{U}_t) \gets \mat{G}_t - \mat{U}_t \,\mathrm{sym}(\mat{U}_t^{\intercal}\mat{G}_t)$\;
    $\tmat{U}_{t+1} \gets \mat{U}_t + \eta_U \mathrm{grad}f(\mat{U}_t)$\;
    $\mat{U}_{t+1} \gets \mathrm{qf}(\tmat{U}_{t+1})$\;
    $\mat{S}_{t+1}  \gets \frac{1}{b}\mat{U}_{t+1}^{\intercal} \mat{A}_t\mat{A}_t^{\intercal}\mat{U}_{t+1}$\;
    Eigendecompose  $\mat{S}_{t+1} = \mat{Q}_{t+1} \mat{D}_{t+1}
    \mat{Q}_{t+1}^{\intercal}$\;
    $\mat{U}_{t+1} \gets \mat{U}_{t+1}\mat{Q}_{t+1}$\;
    $\mat{\Lambda}_{t+1} \gets \beta_\Lambda \mat{\Lambda}_t +
    (1-\beta_\Lambda)\mat{D}_{t+1}$\;
    $T_{t+1} \leftarrow (n_l-r) + \sum_{i=1}^{r} (\lambda_{i,t+1}+\rho)^{-\alpha}$\;
    $s_{t+1} \leftarrow \sqrt{\frac{n_l}{T_{t+1}}}$\; 
    Sample $\mat{R}_t \sim \mathcal{MN}(\mat{0}, \mat{I}_{m_l}, \mat{I}_{n_l})$ \;
    $\begin{aligned}
    \mat{\Delta}_t \gets s_{t+1}\big[&\mat{R}_t + \mat{R}_t \mat{U}_{t+1}\big(
    (\mat{\Lambda}_{t+1}+\rho \mat{I}_r)^{-\alpha/2}\\
    &-\mat{I}_r\big)\mat{U}_{t+1}^{\intercal}\big]
    \end{aligned}$\; 
    $\mathcal{L}_t^+ \gets \ell(\mat{W}_t+\varepsilon \mat{\Delta}_t;\xi_t)$\,,
    $\mathcal{L}_t^- \gets \ell(\mat{W}_t-\varepsilon \mat{\Delta}_t;\xi_t) $\;
    $\hmat{G}_t  \gets  \dfrac{\mathcal{L}_t^+-\mathcal{L}_t^-}{2\varepsilon} \mat{\Delta}_t $\;
    $\mat{W}_{t+1} \gets \mat{W}_t - \eta \hmat{G}_t$\;
  }
  % \Return{$\mat{W}_T$}
\end{algorithm}
\theoryadd{The base Gaussian matrix $\mat{R}_t$ in
Algorithm~\ref{alg:pams} is independent of $\xi_t$ given the past.
The perturbation covariance is constructed from $\xi_t$ before
sampling $\mat{R}_t$, corresponding to the current-batch case
in~\eqref{eq:dependent_bound}.}

\subsection{Online Activation Subspace Tracking}
\label{sec:subspace_tracking}
The activation preconditioner $\mat{P}_l$ in
Section~\ref{sec:metric_construction} depends on the leading
eigenspace and eigenvalues of the activation second-moment matrix
$\mat{\Sigma}_{A,l} = \E{\mat{A}_l\mat{A}_l^{\intercal}}$. For large
transformers, explicitly
forming $\mat{\Sigma}_{A,l}\in \R^{n_l\times n_l}$ or computing its
full eigendecomposition is prohibitively expensive.
PAMS therefore tracks only its leading $r$-dimensional subspace using
streaming mini-batch activations. 
For clarity, we omit the layer index $l$ throughout this subsection
and write $n=n_l$.
The leading $r$-dimensional activation subspace can be characterized as the
solution to the PCA problem 
\begin{equation} \label{eq:pca_objective}
  \max_{\mat{U}\in \mathrm{St}(r,n_l)}
  f(\mat{U}) :=\frac{1}{2}\tr(\mat{U}^{\intercal} \mat{\Sigma}_A \mat{U})\,,
\end{equation}
where $\mathrm{St}(r,n_l) = \{ \mat{U}\in\R^{n_l\times
  r}:\mat{U}^{\intercal} \mat{U} = \mat{I}_r\}$ 
denotes the Stiefel manifold.

\paragraph{Subspace Update}
At iteration $t$, let $\mat{A}_t\in \R^{n\times b}$ denote the
mini-batch activation matrix. PAMS does not explicitly form the
corresponding empirical second-moment matrix $\hmat{\Sigma}_{t} =
\frac{1}{b}\mat{A}_t\mat{A}_t^{\intercal}$. Instead, given the current subspace
$\mat{U}_t$, it computes its action on the subspace directly:
\begin{equation} \label{eq:subspace_action}
    \mat{G}_t = \hmat{\Sigma}_t\mat{U}_t = \frac{1}{b}
  \mat{A}_t(\mat{A}_t^{\intercal}\mat{U}_t)\,.
\end{equation}
This requires $\mathcal{O}(nbr)$ computation and avoids constructing an
$n\times n$ covariance matrix.
The resulting stochastic Riemannian gradient of
~\eqref{eq:pca_objective} is
\begin{equation}
  \mathrm{grad}f(\mat{U}_t) = \mat{G}_t - \mat{U}_t \mathrm{sym}\!\left(
    \mat{U}_t^{\intercal}\mat{G}_t\right)\,,
  \label{eq:riemannian_gradient}
\end{equation}
where
$\mathrm{sym}(\mat{X})=\frac{1}{2}(\mat{X}+\mat{X}^{\intercal})$.
PAMS performs a Riemannian gradient ascent step,
% \begin{equation}
$
  \tmat{U}_{t+1}=\mat{U}_t +\eta_U\,\mathrm{grad}f(\mat{U}_t)$,
% \label{eq:subspace_step}
% \end{equation}
where $\eta_U>0$ is the subspace learning rate.
Since this update does not generally preserve the orthonormality
constraint, we project the result back onto the Stiefel manifold using
a QR retraction: 
% \begin{equation}
$ \mat{U}_{t+1} = \mathrm{qf}\!\left(\tmat{U}_{t+1} \right)$,
% \label{eq:qr_retraction}
% \end{equation}
where $\mathrm{qf}(\cdot)$ denotes the orthonormal $\mat{Q}$ factor of
a QR decomposition.
Thus, $\mat{U}_{t+1}$ is the updated activation subspace obtained from
the current mini-batch. The spectrum estimation described next is
performed using this updated subspace.

\paragraph{Online Spectrum Estimation}
Tracking the dominant subspace alone is insufficient to construct the
preconditioner in Section~\ref{sec:metric_construction}, which also
requires the eigenvalues associated with the retained directions.
After obtaining the updated subspace $\mat{U}_{t+1}$, 
PAMS estimates its activation spectrum by projecting the same
mini-batch covariance onto $\mat{U}_{t+1}$:
% \begin{equation}
\[
  \mat{S}_{t+1} =
  \mat{U}_{t+1}^{\intercal}\hmat{\Sigma}_t\mat{U}_{t+1} =\frac{1}{b}
  (\mat{A}_t^{\intercal}\mat{U}_{t+1})^{\intercal}(\mat{A}_t^{\intercal}\mat{U}_{t+1})\in
  \R^{r\times r}\,.
\]
% \label{eq:projected_covariance}
% \end{equation}
Importantly, $\mat{S}_{t+1}$ is only $r\times r$, so the spectral
computation is independent of the full feature dimension $n$.
PAMS eigendecomposes this projected covariance as
% \begin{equation}
\[
  \mat{S}_{t+1} = \mat{Q}_{t+1} \mat{D}_{t+1} \mat{Q}_{t+1}^{\intercal}\,,
\]
% \label{eq:small_eigendecomposition}
% \end{equation}
where the diagonal entries of $\mat{D}_{t+1}$ are sorted in descending
order. The 
columns of $\mat{U}_{t+1}\mat{Q}_{t+1}$ are the corresponding Ritz directions
of the empirical covariance within the updated subspace. PAMS therefore
rotates the basis as $\mat{U}_{t+1}\gets \mat{U}_{t+1}\mat{Q}_{t+1}$,
so that the columns of $\mat{U}_{t+1}$ are aligned with the estimated activation
spectrum.
Finally, because the spectrum estimated from a single mini-batch can
exhibit substantial sampling variability, PAMS maintains an
exponential moving average of the retained eigenvalues: 
% \begin{equation}
\[
  \mat{\Lambda}_{t+1} =\beta_\Lambda\mat{\Lambda}_t +
(1-\beta_\Lambda)\mat{D}_{t+1}\,,
% \label{eq:eigenvalue_ema}
% \end{equation}
\]
where $\beta_\Lambda\in[0,1)$ controls temporal smoothing.
The resulting pair $(\mat{U}_{t+1},\mat{\Lambda}_{t+1})$
therefore contains the low-rank activation geometry used by PAMS at
the next iteration. Specifically, $\mat{U}_{t+1}$ specifies the
retained activation directions, while $\mat{\Lambda}_{t+1}$ specifies
the corresponding spectral weights used in~\eqref{eq:ai_preconditioner}.

% The tracker therefore requires only $\mathcal{O}(nr)$ storage for the
% subspace and $\mathcal{O}(r)$ storage for the retained spectrum.
% Its dominant computation is the covariance-action operation
% in~\eqref{eq:subspace_action}, which costs $\mathcal{O}(nbr)$, while
% the only eigendecomposition is that of the $r\times r$ matrix
% $\mat{S}_t$, costing $\mathcal{O}(r^{3})$. hence the overall per-layer
% tracking cost is $\mathcal{O}(nbr+r^{3})$, with no dependence on the
% square of the feature dimension $n$. This makes the procedure
% practical for the large feature dimensions encountered in transformer
% layers. 

\paragraph{Computational considerations}
PAMS never forms the $n_l\times n_l$ activation second-moment matrix. The
subspace update requires only the covariance action
$\mat{A}_t(\mat{A}_t^{\intercal}\mat{U}_t)$,
with computational cost $\mathcal{O}(nbr)$, while the projected
spectrum requires an $r\times r$ eigendecomposition with cost
$\mathcal{O}(r^{3})$. The persistent state consists of 
the $n\times r$ subspace basis and $r$ retained eigenvalues, requiring
$\mathcal{O}(nr)$ memory.
Thus, PAMS reduces the memory required for activation-geometry
tracking from $\mathcal{O}(n^{2})$ to $\mathcal{O}(nr)$. When $r\ll
n$, this makes online tracking practical for the large feature
dimensions of transformer layers. The procedure is also fully
compatible with the ZO setting: it relies exclusively on forward-pass
activations and requires no backpropagated gradients, Hessian
information, or Fisher-vector products.
In summary, PAMS maintains a compact, continuously updated
representation of activation geometry. Each mini-batch updates the
dominant subspace, estimates the spectrum within that subspace, and
supplies the low-rank geometric information needed to construct the
activation-informed ZO perturbations.

%%% Local Variables:
%%% mode: latex
%%% TeX-master: "../main"
%%% End:

\section{Convergence Analysis}
\label{sec:analysis}
This section analyzes the convergence of PAMS under standard smoothness
assumptions.
The analysis characterizes the effect of the activation-informed
perturbation geometry through the condition number induced by the
bounded activation spectrum. Due to space constraints, we provide
proof sketches here and completed proofs are provided in the
Appendix~\ref{app:proofs}.
\begin{theoryaddition}
Let $\mathcal{L}(\vec{w})=\E[\xi]{\ell(\vec{w};\xi)}$, and let
$\mathcal{F}_t$ contain the history before drawing the fresh loss batch
$\xi_t$. We write $\E[t]{\cdot}=\E{\cdot\mid\mathcal{F}_t}$.
The iterate $\vec{w}_t$ is $\mathcal{F}_t$-measurable, with fixed
$\vec{w}_0$. The covariance may use the history and $\xi_t$, but is
fixed before drawing $\vec{z}_t\sim\Norm{\vec{0},\mat{I}_d}$
independently of $\xi_t$ given $\mathcal{F}_t$. We set
$\vec{u}_t=\mat{C}_t^{1/2}\vec{z}_t$ and use
\begin{align}
 \hvec{g}_t
 &=\frac{\ell(\vec{w}_t+\varepsilon\vec{u}_t;\xi_t)
          -\ell(\vec{w}_t-\varepsilon\vec{u}_t;\xi_t)}{2\varepsilon}
          \vec{u}_t, \nonumber\\
 \vec{w}_{t+1}&=\vec{w}_t-\eta\hvec{g}_t,
 \qquad \varepsilon>0.
 \label{eq:minibatch_estimator}
\end{align}
Both evaluations use the same batch and loss randomness.
All random quantities are measurable, and smoothing derivatives hold
the covariance fixed.
\end{theoryaddition}
We make the following assumptions.
\begin{enumerate}[label=\textbf{(A\arabic*)},leftmargin=*]
\item \textbf{Lower-bounded objective.} There exists $\mathcal{L}^{\star}>-\infty$
  such that $\mathcal{L}(\vec{w})\geq \mathcal{L}^{\star}$, \theorydel{for} $\forall
  \vec{w}$. \label{assump:lower_bounded}
% \item \textbf{Smoothness.} The objective $\mathcal{L}$ is
%   $M$-smooth \theoryadd{for some $M>0$}; that is $\norm{\grad{\mathcal{L}}(\vec{x}) -
%     \grad{\mathcal{L}}(\vec{y})} \leq M\norm{\vec{x}-\vec{y}}$,
%   $\forall \vec{x}, \vec{y}$.
\item \textbf{Sample-loss smoothness.} Almost surely, $\ell(\cdot;\xi)$
  is continuously differentiable and 
  has an $M$-Lipschitz gradient:
  $\norm{\grad\ell(\vec{x};\xi) - \grad{\ell}(\vec{y};\xi)}
  \leq
  M\norm{\vec{x}-\vec{y}}$, $\forall \vec{x}, \vec{y}\in \R^{d}$.
  Consequently, % the population objective
  the population loss $\mathcal{L}$ is $M$-smooth.
  \label{assump:smoothness}
\item % Uniformly bounded PAMS geometry. There exists constants $0 <
  % c_{\min} \leq c_{\max} <\infty$ such that, $c_{\min}\mat{I}\preceq
  % \mat{C}_t \preceq c_{\max}\mat{I}$.
  \textbf{Bounded activation spectrum.} For every layer $l$, iteration
  $t$, and retained eigenvalue $i\leq r$, $0\leq \lambda_{l, i, t}\leq
  \Lambda_{\max}$ \theoryadd{almost surely}, for some finite constant $\Lambda_{\max}$.  
  \label{assump:bounded_geometry}
\item \begin{theoryaddition}
  \textbf{Mini-batch oracle.} The fresh mini-batch has the same sampling law
  conditionally on $\mathcal{F}_t$.
  For every $\vec{x}$,
  \begin{align*}
    \E[\xi]{\grad\ell(\vec{x};\xi)}&=\grad\mathcal{L}(\vec{x})\,,\\
    \E[\xi]{\norm{\grad\ell(\vec{x};\xi)-\grad\mathcal{L}(\vec{x})}^{2}}
      &\leq\sigma_b^{2}.
  \end{align*}
  The loss is integrable at the origin, $\E[\xi]{\abs{\ell(\vec{0};\xi)}}<\infty$.
  For a batch of $b$ independent examples with individual gradient
  variance bounded by $\sigma_1^2$, $\sigma_b^2\leq\sigma_1^2/b$.
  \label{assump:stochastic_oracle}
  \end{theoryaddition}
\end{enumerate}
The bounded activation spectrum in~\ref{assump:bounded_geometry}, 
together with the damping and trace normalization in Section~\ref{sec:method},
implies the following useful spectral bounds. 
Define $q_{\min} = \min\{1,
(\Lambda_{\max}+\rho)^{-\alpha}\}$, $q_{\max}= \max\{1,
\rho^{-\alpha}\}$, and $\kappa_P = \frac{q_{\max}}{q_{\min}}$.
The eigenvalues of the unnormalized preconditioner $\mat{P}_{l,t}$ lie
in $[q_{\min}, q_{\max}]$. Since trace normalization divides these
eigenvalues by their average, which also lies in this interval, every
eigenvalue of the normalized preconditioner $\bmat{P}_{l,t}$ satisfies 
\begin{equation}
  \label{eq:spectral_bounds}
  \kappa_P^{-1}\leq \lambda_i(\bmat{P}_{l,t})\leq \kappa_P\,.
\end{equation}
Consequently, the full perturbation covariance
\[
  \mat{C}_t = \blkdiag(\bmat{P}_{1,t}\otimes \mat{I}_{m_1}, \ldots,
  \bmat{P}_{L,t}\otimes \mat{I}_{m_L})
\]
satisfies
$
  \kappa_P^{-1}\mat{I}_d\preceq \mat{C}_t\preceq \kappa_P\mat{I}_d$.
Moreover, the trace normalization gives $\tr(\mat{C}_t) =
\sum_{l=1}^{L}m_ln_l=d$.
Thus, PAMS preserves the total perturbation variance while allowing
the perturbation covariance to remain anisotropic within each layer. 

The convergence analysis relies on two properties of the antithetic
ZO estimator. First, \theorydel{because the estimator targets the
gradient of the Gaussian-smoothed objective, we quantify the bias} \theoryadd{we use the
Gaussian-smoothed objective to quantify the bias in its
covariance-preconditioned gradient. Specifically, we bound the error}
introduced by the finite smoothing radius $\varepsilon$. Second, we bound the
second moment of the estimator, which controls the error introduced by
the stochastic perturbation in the smoothness-based descent
argument. The following two lemmas establish these properties under
the covariance bounds implied by~\ref{assump:bounded_geometry}
\theoryadd{ and the mini-batch oracle in~\ref{assump:stochastic_oracle}}.

\begin{lemma}[Smoothing bias] \label{lem:smoothing_bias}
  Let $\vec{g}_t = \grad{\mathcal{L}}(\vec{w}_t)$ and let
  $\mathcal{L}_{\varepsilon, \mat{C}_t}(\vec{w}) =
  \E[\vec{u}_t]{\mathcal{L}(\vec{w} + \varepsilon\vec{u}_t)}$, where
  $\vec{u}_t\sim\Norm{\vec{0}, \mat{C}_t}$. If $\tr(\mat{C}_t) =
  d$\theoryadd{, with $\mat{C}_t$ held fixed in this Gaussian expectation}, then
  \[
    \norm{\grad{\mathcal{L}}_{\varepsilon,  \mat{C}_t}(\vec{w}_t) -
      \vec{g}_t}\leq M\varepsilon \sqrt{d}\,.
  \]
\end{lemma}
\begin{proof}
  By $M$-smoothness,
  \begin{align*}
    \norm{\grad{\mathcal{L}}_{\varepsilon,  \mat{C}_t}(\vec{w}_t) -
    \vec{g}_t}
    &=\norm{\E{\grad{\mathcal{L}(\vec{w}_t+\varepsilon\vec{u}_t)} -
      \grad{\mathcal{L}}(\vec{w}_t)}}\\
    &\leq M\varepsilon\E{\norm{\vec{u}_t}} 
    \leq M\varepsilon\sqrt{\tr(\mat{C}_t)}\,.
  \end{align*}
  where the last inequality is obtained by applying Jensen's inequality.
  Because of the trace normalization, we have
  \[
    \tr(\mat{C}_t) = \sum_{l}m_l \tr(\bmat{P}_{l,t}) = \sum_{l}m_ln_l = d\,.
  \]
  Thus, $\norm{\grad{\mathcal{L}}_{\varepsilon,  \mat{C}_t}(\vec{w}_t) -
      \vec{g}_t}\leq M\varepsilon\sqrt{d}$.
\end{proof}
The smoothing bias alone does not characterize the stochasticity of
the ZO update. We therefore next bound the second moment of
the antithetic estimator. Under the spectral bounds
in~\eqref{eq:spectral_bounds}, the estimator satisfies the following
bound.
\begin{lemma}[Second moment bound] \label{lem:second_moment}
  Let $\vec{g}_t = \grad{\mathcal{L}}(\vec{w}_t)$ with
  $\vec{u}_t\theoryadd{\mid(\mathcal{F}_t,\xi_t)}\sim\Norm{\vec{0}, \mat{C}_t}$, and let 
  $\hvec{g}_t$ denote the antithetic PAMS estimator defined in
  Section~\ref{sec:perturbation_sampling}
  % \theoryadd{, evaluated on a shared mini-batch as
    % in~\eqref{eq:minibatch_estimator}}
  . Under~\ref{assump:smoothness}
  \theoryadd{-~\ref{assump:stochastic_oracle}},   
  \begin{align} \label{eq:second_moment_bound}
    \Ev_{\theoryadd{t}}\norm{\hvec{g}_t}^{2}&\leq 2\kappa_P(d +
    2\kappa_P)\theoryadd{\big(}\norm{\vec{g}_t}^{2}\theoryadd{+\sigma_b^2\big)}\nonumber\\
    &\qquad+\frac{15}{2}M^{2}d^{3}\varepsilon^{2}\,.
  \end{align}
\end{lemma}
\begin{proof}[Proof Sketch]
  \theoryadd{Conditioned on $(\mathcal{F}_t,\xi_t)$, % ; expectations over
    % $\vec{u}_t$ below use the resulting fixed covariance.
    the covariant $\mat{C}_t$ is fixed.
}
  Let $s_t = \frac{\theorydelmath{\mathcal{L}}\theoryadd{\ell}(\vec{w}_t+\varepsilon\vec{u}_t\theoryadd{;\xi_t}) -
  \theorydelmath{\mathcal{L}(\vec{w}_-+\varepsilon\vec{u}_t)}\theoryadd{\ell(\vec{w}_t-\varepsilon\vec{u}_t;\xi_t)}}{2\varepsilon}$ and $s_t
= \vec{g}_{t\theoryadd{,\xi}}^{\intercal}\vec{u}_t + \gamma_t$, where $\vec{g}_{t\theoryadd{,\xi}} =
\grad{\theorydelmath{\mathcal{L}}\theoryadd{\ell}}(\vec{w}_t\theoryadd{;\xi_t})$.
Under~\theorydel{\ref{assump:bounded_geometry}}\theoryadd{\ref{assump:smoothness}},
the gradient is $M$-Lipschitz, and the fundamental theorem of calculus
gives $\abs{\gamma_t}\leq \frac{M\varepsilon}{2}\norm{\vec{u}_t}^{2}$.
Hence, 
\begin{align*}
  \norm{\hvec{g}_t }^{2}
  &\leq  2(\vec{g}_{t\theoryadd{,\xi}}^{\intercal}\vec{u}_t)^{2} \norm{\vec{u}_t}^{2}+
    2\gamma_t^{2}\norm{\vec{u}_t}^{2}\,.
\end{align*}
% For the first term, the Gaussian \theorydel{forth}\theoryadd{fourth}-moment identity gives
% \begin{align*}
%   \E{(\vec{g}_{t\theoryadd{,\xi}}^{\intercal}\vec{u}_t)^{2}\norm{\vec{u}_t}^{2}}
%   &=(\vec{g}_{t\theoryadd{,\xi}}^{\intercal}\mat{C}_t\vec{g}_{t\theoryadd{,\xi}})\tr(\mat{C}_t)+2\vec{g}_{t\theoryadd{,\xi}}^{\intercal}\mat{C}_t^{2}\vec{g}_{t\theoryadd{,\xi}}\\
%   \intertext{Since $\mat{C}_t\preceq \kappa_P\mat{I}_d$ and
%   $\tr(\mat{C}_t) = d$,}
%   \E{(\vec{g}_{t\theoryadd{,\xi}}^{\intercal}\vec{u}_t)^{2}\norm{\vec{u}_t}^{2}}
%   &\leq(d \kappa_P+ 2\kappa_P^{2})\norm{\vec{g}_{t\theoryadd{,\xi}}}^{2}\,.
% \end{align*}
% For the remainder term,
% \begin{align*}
%   2\gamma_t^{2}\norm{\vec{u}_t}^{2}
%   &\leq 2 \frac{M^{2}\varepsilon^{2}}{4}\norm{\vec{u}_t}^{6} =
%     \frac{M^{2}\varepsilon^{2}}{2}\norm{\vec{u}_t}^{6}
      % \end{align*}
Using Gaussian fourth moments, $\mat{C}_t\preceq \kappa_P\mat{I}_d$,
and $\tr(\mat{C}_t)=d$, 
\[
  \E[\vec{u}_t]{(\vec{g}_{t\theoryadd{,\xi}}^{\intercal}\vec{u}_t)^{2}
    \norm{\vec{u}_t}^{2}}\leq
  \kappa_P(d+2\kappa_P)\norm{\vec{g}_{t,\xi}}^{2}\,.
\]
Moreover, the sixth Gaussian-moment bound gives 
\begin{align*}
  \E{2\gamma_t^{2}\norm{\vec{u}_t}^{2}}
  &\leq \frac{15M^{2}\varepsilon^{2}}{2}\tr(\mat{C}_t)^{3} =
    \frac{15M^{2}\varepsilon^{2}d^{3}}{2}\,.
\end{align*}
Combining these bounds, we get
\[
  \E[t]{\norm{\hvec{g}_t}^{2}} \leq 2\kappa_P(d +
  2\kappa_P)\norm{\vec{g}_{t\theoryadd{,\xi}}}^{2} + \frac{15M^{2}\varepsilon^{2}d^{3}}{2}\,.
\]
\begin{theoryaddition}
Finally, taking expectation over $\xi_t$, unbiasedness and the variance bound imply
\[
 \E[t]{\norm{\vec{g}_{t,\xi}}^2}
 =\norm{\vec{g}_t}^2+\E[t]{\norm{\vec{g}_{t,\xi}-\vec{g}_t}^2}
 \leq\norm{\vec{g}_t}^2+\sigma_b^2.
\]
% This proves~\eqref{eq:second_moment_bound}, including when
% $\mat{C}_t$ uses $\xi_t$. The Gaussian moment calculations are given in
      %       Appendix~\ref{app:second_moment}.
The result holds even when $\mat{C}_t$ depends on $\xi_t$, since all
Gaussian moment bounds are established conditionally on
$(\mathcal{F}_t, \xi_t)$ and hold uniformly almost surely.
\end{theoryaddition}
\end{proof}
Lemmas~\ref{lem:smoothing_bias}
and~\ref{lem:second_moment} provide the bias and second-moment controls
needed to establish the convergence result below.
\begin{theorem}[Convergence of PAMS] \label{thm:convergence}
  Under Assumptions~\ref{assump:lower_bounded}
  -~\theorydel{\ref{assump:bounded_geometry}}\theoryadd{\ref{assump:stochastic_oracle}},
  \theoryadd{suppose $\mat{C}_t$ is $\mathcal{F}_t$-measurable, i.e.,
    selected before the fresh loss batch.} Let PAMS
  use a constant step 
  size $\eta_t = \eta$ satisfying $\theoryadd{0<}\eta\leq
  \frac{1}{4M\kappa_P^{2}(d+2\kappa_P)}$. Then the iterates generated
  by PAMS satisfy
  \begin{align*}
    \frac{1}{T}\sum_{t=0}^{T-1}\Ev\norm{\grad{\mathcal{L}}(\vec{w}_t)}^{\theoryadd{2}}
    &\leq \frac{4\kappa_P(\mathcal{L}(\vec{w}_0) - \mathcal{L}^{\star})}{\eta
      T} +
      2M^{2}d\varepsilon^{2}\kappa_P^{\theorydelmath{3}\theoryadd{4}}\\
    &\qquad
      + 15M^{3}d^{3}\varepsilon^{2}\eta\kappa_P\\
    &\qquad\theoryadd{+4M\eta\kappa_P^2(d+2\kappa_P)\sigma_b^2}\,.
  \end{align*}
\end{theorem}
\begin{proof}[Proof Sketch]
  Let $\vec{b}_t = \grad{\mathcal{L}}_{\varepsilon,
    \mat{C}_t}(\vec{w}_t) - \vec{g}_t$.
  Since $\mat{C}_t$ and $\vec{w}_t$ are $\mathcal{F}_t$-measurable,
  applying Lemma~\ref{lem:zo_precond} conditionally and using the
  unbiasedness of the fresh mini-batch gives 
  $\E[t]{\hvec{g}_t} = \mat{C}_t(\vec{g}_t+\vec{b}_t)$. 
  % \theoryadd{Here the conditional mean averages over both the batch
  % and the Gaussian draw. The sampling assumption permits replacing
  % the sample loss by $\mathcal{L}$ before applying Lemma~\ref{lem:zo_precond}.}
  By the $M$-smoothness of $\mathcal{L}$, we have
  \begin{align}
    \mathcal{L}(\vec{w}_{t+1})
    &\leq \mathcal{L}(\vec{w}_t) - \eta_t\ip{\vec{g}_t, \hvec{g}_t} +
      \frac{M\eta_t^{2}}{2}\norm{\hvec{g}_t}^{2} \nonumber \\
    \intertext{Taking conditional expectation and applying Lemma~\ref{lem:zo_precond},}
    \E[t]{\mathcal{L}(\vec{w}_{t+1})}
    % &\leq \mathcal{L}(\vec{w}_t) - \eta_t\ip{\vec{g}_t, \E[t]{\hvec{g}_t}}
    %   + \frac{M\eta_t^{2}}{2}\E[t]{\norm{\hvec{g}_t}^{2}} \nonumber \\
    % &= \mathcal{L}(\vec{w}_t) - \eta_t\ip{\vec{g}_t,
    %   \mat{C}_t\grad{\mathcal{L}}_{\varepsilon, \mat{C}_t}(\vec{w}_t)}
    %   \nonumber \\
    % &\qquad +
    % \frac{M\eta_t^{2}}{2}\E[\theoryadd{t}]{\norm{\hvec{g}_t}^{2}}\,. \label{eq:exp_descent}
&\leq \mathcal{L}(\vec{w}_t) -
\eta_t\vec{g}_t^{\intercal}\mat{C}_t(\vec{g}_t+\vec{b}_t) \nonumber \\
&\qquad +\frac{M\eta_t^{2}}{2}\E[\theoryadd{t}]{\norm{\hvec{g}_t}^{2}}\,. \label{eq:exp_descent}
  \end{align}
  %  Then,
  % \[
  %   \ip{\vec{g}_t, \mat{C}_t\grad{\mathcal{L}}_{\varepsilon,
  %       \mat{C}_t}(\vec{w}_t)}
  %   = \vec{g}^{\intercal}\mat{C}_t \vec{g}_t + \vec{g}_t^{\intercal}\mat{C}_t\vec{b}_t\,.
  % \]
  By Assumption~\ref{assump:bounded_geometry},
  $\vec{g}_t^{\intercal}\mat{C}_t\vec{g}_t\geq
  \kappa_P^{-1}\norm{\vec{g}_t}^{2}$. Moreover,
  $\abs{\vec{g}_t^{\intercal}\mat{C}_t\vec{b}_t} \leq
  \kappa_P\norm{\vec{g}_t}\norm{\vec{b}_t}$. Using Young's inequality,
  $
    \kappa_P\norm{\vec{g}_t}\norm{\vec{b}_t}\leq
    \frac{1}{2\kappa_P}\norm{\vec{g}_t}^{2}+\frac{\kappa_P^{3}}{2}\norm{\vec{b}_t}^{2}\,.
  $
  Thus,
  \begin{equation}
    \label{eq:gradip_bound}
    \ip{\vec{g}_t, \mat{C}_t\grad{\mathcal{L}}_{\varepsilon,
        \mat{C}_t}(\vec{w}_t)}
    \geq
    \frac{1}{2\kappa_P}\norm{\vec{g}_t}^{2}-\frac{\kappa_P^{3}M^{2}d\varepsilon^{2}}{2}\,.
  \end{equation}
  Substituting the gradient second-moment bound
  in~\eqref{eq:second_moment_bound} and~\eqref{eq:gradip_bound}
  into~\eqref{eq:exp_descent} yields
  \begin{align*}
    \E[t]{\mathcal{L}(\vec{w}_{t+1})}
    % &\leq \mathcal{L}(\vec{w}_t) +
    %   \eta_t\left(\frac{\kappa_P^{3}M^{2}d\varepsilon^{2}}{2}-\frac{1}{2\kappa_P}\norm{\vec{g}_t}^{2}\right)\\  
    % &\qquad
    %   +\frac{M\eta_t^{2}}{2}\left(2\kappa_P(d + 2\kappa_P)\norm{\vec{g}_t}^{2}+\frac{15}{2}M^{2}d^{3}\varepsilon^{2}\right)\\
    &\leq \mathcal{L}(\vec{w}_t)\\
    &\qquad-\frac{\eta_t}{2}\left[\kappa_P^{-1} -
      2M\eta_t\kappa_P(d+2\kappa_P)\right]\norm{\vec{g}_t}^{2} \\
    &\qquad +
      \left[\frac{\kappa_P^{3}M^{2}d\varepsilon^{2}\eta_t}{2}
      +\frac{15M^3d^{3}\varepsilon^{2}\eta_t^{2}}{4}\right]\\
    &\qquad\theoryadd{+M\eta_t^2\kappa_P(d+2\kappa_P)\sigma_b^2}\,.
  \end{align*}
  If $\eta_t \leq \frac{1}{4M\kappa_P^{2}(d+2\kappa_P)}$, then $\kappa_P^{-1} -
  2M\kappa_P(d+2\kappa_P) \eta_t\geq \frac{1}{2\kappa_P}$.
  \begin{align*}
    &\E[t]{\mathcal{L}(\vec{w}_{t+1})}\\
    &\leq \mathcal{L}(\vec{w}_t) 
      -\frac{\eta_t}{4\kappa_P}\norm{\vec{g}_t}^{2} 
      +
      \left[\frac{\kappa_P^{3}M^{2}d\varepsilon^{2}\eta_t}{2}
      +\frac{15M^3d^{3}\varepsilon^{2}\eta_t^{2}}{4}\right]\\
    &\qquad\theoryadd{+M\eta_t^2\kappa_P(d+2\kappa_P)\sigma_b^2}\,.
  \end{align*}
  Let $\eta_t = \eta$. Taking total expectations and summing from
  $t=0$ to $T-1$, we get
  $\frac{\eta }{4\kappa_P}\sum_{t=0}^{T-1}\E{\norm{\vec{g}_t}^{2}}
  \leq \mathcal{L}(\vec{w}_0) - \theorydelmath{\mathcal{L}_*}\theoryadd{\mathcal{L}^{\star}}
  +
      TM^{2}d\varepsilon^{2}\eta\left[\frac{\kappa_P^{3}}{2} +
      \frac{15Md^{2}\eta}{4}\right]
    \theoryadd{+TM\eta^2\kappa_P(d+2\kappa_P)\sigma_b^2}$.
    Dividing by $T\eta /(4\kappa_P)$ gives
  \begin{align*}
    % \frac{\eta }{4\kappa_P}\sum_{t=0}^{T-1}\E{\norm{\vec{g}_t}^{2}}
    % &\leq \mathcal{L}(\vec{w}_0) - \theorydelmath{\mathcal{L}_*}\theoryadd{\mathcal{L}^{\star}}\\
    % &\qquad
    %   +
    %   TM^{2}d\varepsilon^{2}\eta\left[\frac{\kappa_P^{3}}{2} +
    %   \frac{15Md^{2}\eta}{4}\right]\,. \\
    % \intertext{Dividing by $T\eta /(4\kappa_P)$ gives}
    \frac{1}{T}\sum_{t=0}^{T-1}\E{\norm{\vec{g}_t}^{2}}
    &\leq\frac{4\kappa_P(\mathcal{L}(\vec{w}_0) - \theorydelmath{\mathcal{L}_*}\theoryadd{\mathcal{L}^{\star}})}{\eta T}
      + 2M^{2}d \varepsilon^{2}\kappa_P^{\theorydelmath{3}\theoryadd{4}} \\
    &\qquad +
      15\kappa_PM^{3}d^{3}\varepsilon^{2}\eta\\
    &\qquad\theoryadd{+4M\eta\kappa_P^2(d+2\kappa_P)\sigma_b^2}\,.
  \end{align*}
\theoryadd{Details of the complete conditional-expectation and
summation argument are provided in Appendix~\ref{app:convergence}.}
\end{proof}
For a constant step size $\eta$ and smoothing radius $\varepsilon$,
PAMS achieves an $\mathcal{O}((\eta
T)^{-1})+\mathcal{O}(\varepsilon^{2})
+\mathcal{O}(\eta\varepsilon^{2})\theoryadd{+\mathcal{O}(\eta\sigma_b^2)}$ bound on the average \theoryadd{expected} squared
gradient norm. Choosing \theoryadd{an admissible}
$\eta=\theorydelmath{\mathcal{O}}\theoryadd{\Theta}(T^{-1/2})$ yields the
standard $\mathcal{O}(T^{-1/2})+\mathcal{O}(\varepsilon^{2})$,
corresponding to an $\mathcal{O}(\varepsilon^{2})$-stationary
neighborhood of the population objective for fixed $\varepsilon$.
% Thus, as $T\to \infty$,
% \theorydel{PAMS converges to an
% $\mathcal{O}(\varepsilon^{2})$-stationary neighborhood of the original
% objective}
% \theoryadd{the asymptotic upper bound on this average is
% $\mathcal{O}(\varepsilon^{2})$ with this step-size schedule}
% for fixed smoothing radius $\varepsilon$.
Decreasing $\varepsilon$ appropriately with $T$ drives the bound to zero.
\begin{theoryaddition}
With fixed $\eta$ and $\varepsilon$, the bound retains both
smoothing and mini-batch errors. These rates assume that the
covariance is
selected before the fresh loss batch.

\paragraph{Geometry from the current mini-batch.}
When $\mat{C}_t$ is constructed from $\xi_t$, as in
Algorithm~\ref{alg:pams}, define
\begin{align}
 \bmat{C}_t&=\E[t]{\mat{C}_t},\qquad
 \vec{g}_t=\grad\mathcal{L}(\vec{w}_t),\nonumber\\
 \vec{r}_t&=\E[t]{\mat{C}_t(\grad\ell(\vec{w}_t;\xi_t)-\vec{g}_t)},
 \label{eq:dependence_residual}\\
 R_T&=\frac1T\sum_{t=0}^{T-1}\E{\norm{\vec{r}_t}^2}.\nonumber
\end{align}
Under
Assumptions~\ref{assump:lower_bounded}--\ref{assump:stochastic_oracle}
and the same step-size restriction,
we obtain the following bound:
\begin{align}
 \frac1T\sum_{t=0}^{T-1}\E{\norm{\vec{g}_t}^2}
 &\leq\frac{4\kappa_P(\mathcal{L}(\vec{w}_0)-\mathcal{L}^{\star})}{\eta T}
 \nonumber\\
 &\quad+4M^2d\varepsilon^2\kappa_P^4+4\kappa_P^2R_T\nonumber\\
 &\quad+4M\eta\kappa_P^2(d+2\kappa_P)\sigma_b^2\nonumber\\
 &\quad+15M^3d^3\varepsilon^2\eta\kappa_P.
 \label{eq:dependent_bound}
\end{align}
The residual $R_T$ measures covariance--batch dependence and
is zero under the sampling condition of Theorem~\ref{thm:convergence}.
For
current-batch geometry, making this upper bound vanish also
requires $R_T\to0$; decreasing $\eta$ and $\varepsilon$ alone does
not ensure this. Exact objective evaluations recover
Theorem~\ref{thm:convergence} with $\sigma_b=0$.
\end{theoryaddition}

%%% Local Variables:
%%% mode: latex
%%% TeX-master: "../main"
%%% End:

\section{Experiments}
\label{sec:experiment}
\subsection{Experimental Setup}
\label{sec:exp_setup}
We evaluate PAMS across three model scales and two transformer
architectures: RoBERTa-large~\cite{Liu2019RoBERTaAR}, OPT-2.7B, and
OPT-13B~\cite{Zhang2022OPTOP}.
% This setup is designed to assess
% whether activation-informed ZO preconditioning generalizes across
% encoder-based and autoregressive language models and whether its
% effectiveness persists as the number of optimized parameters
% increases.
We consider full-parameter fine-tuning on a diverse collection of
natural language understanding tasks from the GLUE and SuperGLUE benchmarks.
We compare PAMS with three representative ZO fine-tuning
methods. \textsc{MeZO}~\cite{malladi2023finetuning} serves as a 
canonical memory-efficient ZO baseline based on isotropic random
perturbations. \textsc{HiZOO}~\cite{zhao2025secondorder} incorporates
diagonal Hessian information to account for heterogeneous curvature,
while \textsc{LOZO}~\cite{chen2025enhancing} exploits low-rank
structure by performing ZO optimization over low-dimensional
subspaces. These baselines represent three complementary approaches
to improve ZO optimization: isotropic perturbation, 
curvature-informed preconditioning, and low-rank subspace
optimization. Together, they provide a meaningful basis for
evaluating the activation-informed perturbation geometry of PAMS.
All experiments were conducted on a cluster equipped with a single
NVIDIA A100 or H100 GPU, and results are reported as mean accuracy with
the sample standard deviation across independent random seeds.

\subsection{Masked Language Models}
\label{sec:main_results}
We first evaluate PAMS on RoBERTa-large, a 355M-parameter encoder-only
transformer, across six natural language understanding
tasks. Table~\ref{tab:roberta_results} reports the test accuracy as percentages, 
averaged over five random seeds, together with the corresponding
standard deviations.
PAMS achieves the highest mean performance on four of the
six tasks and obtains the best overall average score of $\mathbf{69.50\%}$.
Specifically, PAMS achieves the highest mean accuracy on SST-2
($91.78\%$), SST-5 ($47.32\%$), MNLI ($62.80\%$), and RTE
($67.29\%$). On SNLI, HiZOO achieves the highest accuracy of
$71.12\%$, while PAMS remains competitive at $70.76\%$. On TREC, 
LOZO achieves the highest accuracy ($79.32\%$), compared with
$77.08\%$ for PAMS.
PAMS also achieves substantial improvements over the strongest
baseline on several tasks. In particular, PAMS improves upon the
strongest competing method by $4.83$ percentage points on RTE and 
$1.38$ percentage points on SST-5.
These gains, together with PAMS's strong performance on SST-2 and MNLI,
lead to the highest average score across all six tasks. Importantly,
the improvements are observed across multiple tasks rather than being
driven by a single benchmark. 
Overall, these results provide empirical evidence that incorporating
forward-pass activation geometry into zeroth-order exploration can
improve both the effectiveness and robustness of ZO
fine-tuning. In particular, the consistent gains across tasks with
different characteristics suggest that the benefit of PAMS is not tied
to a single task or dataset, but arises from its ability to adapt the
perturbation distribution to the structure of the model's intermediate
representations. 

\begin{table*}[htbp]
\centering
\caption{Results on RoBERTa-large. Reported values are the task
  evaluation metrics in percentage points. Each entry is reported as
  mean $\pm$ sample standard deviation across 5 random seeds. Best results are shown in bold.}
\label{tab:roberta_results}
\begin{tabular}{lcccccccc}
\toprule
Method
& SST-2
& SST-5
& SNLI
& MNLI
& RTE
& TREC
& Avg.\\
  \midrule
  MeZO 
  & \(90.30{\pm}1.0\)
  & \(45.94{\pm}2.5\)
  & \(67.96{\pm}7.1\)
  & \(58.16{\pm}3.2\)
  & \(62.46{\pm}5.0\)
& \(73.64{\pm}4.6\)
& $66.41$ \\
LOZO
  & \(90.61{\pm}1.5\)
  & \(45.08{\pm}2.6\)
  & \(69.02{\pm}3.8\)
  & \(58.64{\pm}2.4\)
  & \(62.16{\pm}3.0\)
& \(\mathbf{79.32}{\pm}3.0\)
  & $67.47$\\
HiZOO
  & \(89.56{\pm}2.0\)
  & \(43.88{\pm}2.9\)
  & \(\mathbf{71.12}{\pm}3.9\)
  & \(61.60{\pm}1.8\)
  & \(61.01{\pm}5.8\)
& \(74.00{\pm}3.6\)
& $66.86$\\
\midrule
\textbf{PAMS}
& $\mathbf{91.78}{\pm}1.09$
& $\mathbf{47.32}{\pm}0.82$
& $70.76{\pm}2.56$
& $\mathbf{62.8}{\pm}1.26$
& $\mathbf{67.29}{\pm}4.25$
& $77.08{\pm}5.77$
  & $\mathbf{69.50}$ \\
\bottomrule
\end{tabular}
\end{table*}

% \paragraph{Scaling to OPT models.}
% We next evaluate PAMS on OPT-2.7B and OPT-13B. Scaling from RoBERTa-large to multi-billion-parameter autoregressive models substantially increases the ambient dimension of the ZO problem, providing a stringent test of whether activation-informed perturbations remain effective in the regime where isotropic estimation is particularly challenging.

% Both OPT models are evaluated on the same ten-task suite. We report accuracy for SST-2, RTE, CB, BoolQ, WSC, WiC, MultiRC, COPA, and ReCoRD, and token F1 for SQuAD. Because the task columns do not all represent the same evaluation metric, we report individual task results rather than constructing an official GLUE/SuperGLUE aggregate score.

\begin{table*}[htbp]
\centering
\caption{Results on OPT-2.7B. Values are reported as mean $\pm$ sample standard deviation across seeds. All values are percentages.}
\label{tab:opt27b_results}
\setlength{\tabcolsep}{6pt}

% ---------------- First 6 datasets ----------------
\begin{tabular}{lcccccccc}
\toprule
Method
& SST-2 & RTE & CB & BoolQ & WiC & MultiRC & COPA & Mean (gain \%p)\\
\midrule

MeZO
& $90.23{\pm}0.25$
& $62.77{\pm}0.25$
& $\mathbf{69.77{\pm}0.25}$
& $67.50{\pm}0.30$
% & $\mathbf{61.60{\pm}0.53}$
& $57.60{\pm}0.53$ 
% ---------------------
& $58.77{\pm}0.25$
& $82.50{\pm}0.50$
& $69.87$ $(\,\,\,0.00\%)$\\  
LOZO
& $\mathbf{93.73{\pm}0.31}$
& $57.23{\pm}1.46$
& $67.80{\pm}0.00$
& $67.03{\pm}1.87$
% & $51.87{\pm}11.13$
& $55.23{\pm}0.47$ 
% ---------------------
& $60.40{\pm}0.82$
& $77.00{\pm}1.73$
& $68.34$ $(-1.53\%)$ \\
HiZOO
& $90.95 \pm 0.15$
& $67.10\pm 3.25$
& $66.95\pm 1.72$
& $61.65\pm 3.15$
% & $59.10$
& $50.95 \pm 2.85$ 
% ---------------------
& $46.90 \pm 3.95$
& $78.00 \pm 0.50$  
& $66.07$ $(-3.81\%)$\\
  \midrule
\textbf{PAMS}
& $93.42{\pm}0.63$
& $\mathbf{67.39{\pm}1.16}$
& $69.00{\pm}2.08$
& $\mathbf{69.67{\pm}0.38}$
% & $51.05*{\pm}0.05$
& $\mathbf{58.20{\pm}1.41}$ 
% ---------------------
& $\mathbf{63.23{\pm}1.55}$
& $\mathbf{82.67{\pm}1.53}$
& $\mathbf{71.94}$ $(+2.07\%)$\\
\bottomrule
\end{tabular}

\vspace{0.8em}
% ---------------- Remaining 4 datasets + Mean ----------------
% \begin{tabular}{lccccc}
% \toprule
% Method
% & MultiRC & COPA & ReCoRD & SQuAD & Mean \\
% \midrule

% MeZO
% & $58.77{\pm}0.25$
% & $82.50{\pm}0.50$
% & --
% & --
% & $68.84$ \\

% LOZO
% & $60.40{\pm}0.82$
% & $77.00{\pm}1.73$
% & --
% & --
% & $66.29$ \\

% HiZOO
% & $46.90$
% & $78.00$
% & --
% & --
% & $65.18$ \\

% \textbf{PAMS}
% & $\mathbf{63.23{\pm}1.55}$
% & $\mathbf{82.67{\pm}1.53}$
% & --
% & --
% & $\mathbf{68.96}$ \\

% \bottomrule
% \end{tabular}
\end{table*}
% =========================================================
% OPT-13B RESULTS
% =========================================================
\begin{table*}[htbp]
\centering
\caption{Results on OPT-13B. Values are reported as mean $\pm$
  sample standard deviation across seeds. All values are
  percentages.}
\label{tab:opt13b_results}

\setlength{\tabcolsep}{6pt}

% ---------------- First 6 datasets ----------------
\begin{tabular}{lcccccc}
\toprule
Method
& SST-2 & RTE & CB & BoolQ & WSC & Mean (gain \%p)\\
\midrule

MeZO
& $92.57\pm0.38$ & $67.33\pm2.25$ & $67.20\pm0.85$ & $71.43\pm3.03$ & $51.87\pm12.11$ & $70.08$ $(\,\,\,0.00\%)$\\

LOZO
& $92.60\pm0.86$ & $64.33\pm1.70$ & $69.60\pm1.47$ & $71.07\pm2.63$ &
                                                                      $51.87\pm10.98$ &$69.89$ $(-0.19\%)$  \\

HiZOO
& $86.60\pm 3.7$ & $67.50\pm 1.50$ & $67.45 \pm 0.35$ & $68.9\pm 4.40$
                           & $\mathbf{58.65\pm 1.95} $ & $69.82$ $(-0.26\%)$\\
\midrule
\textbf{PAMS}
& $\mathbf{92.66\pm0.00}$ & $\mathbf{69.80\pm2.18}$ &
                                                      $\mathbf{71.43\pm1.79}$
                   & $\mathbf{74.00\pm2.67}$ & $ 54.81\pm1.67$ &
                                                                 $\mathbf{72.54}$ $(+2.46\%)$\\
\bottomrule
\end{tabular}
\vspace{0.8em}

% ---------------- Remaining 4 datasets + Mean ----------------
\begin{tabular}{lcccccc}
\toprule
Method
& WiC & MultiRC & COPA & ReCoRD & SQuAD & Mean (gain \%p)\\
\midrule

MeZO
&$58.07\pm0.79$ & $64.33\pm1.41$ & $\mathbf{84.67\pm1.25}$ & $\mathbf{81.20\pm
                                                    0.12}$  &
                                                              $83.10\pm
                                                              1.20$ &
                                                                      $74.27$ $(\,\,\,0.00\%)$ \\

LOZO
& $57.07\pm1.07$ & $61.07\pm3.04$ & $83.33\pm0.47$ & $78.20\pm 1.55$ &
                                                                       $83.55\pm 0.68$ & $72.64$ $(-1.63\%)$ \\

HiZOO
&$54.30\pm 5.90$  & $\mathbf{65.80\pm 8.20}$ & $82.50\pm 2.50$ & $80.90\pm
                                                        0.29$ &
                                                                $83.70\pm
                                                                0.58$
                                        &$73.44$ $(-0.83\%)$ \\
  \midrule
\textbf{PAMS}
&$\mathbf{59.77\pm2.90}$  & $65.07\pm3.88$ &
                                                         $\mathbf{84.67\pm1.53}$ & $80.53\pm0.95$ & $\mathbf{84.17\pm1.48}$ & $\mathbf{74.84}$ $(+0.57\%)$ \\
\bottomrule
\end{tabular}

\end{table*}
\subsection{Autoregressive Language Models}
We next evaluate PAMS on OPT-2.7B and OPT-13B to examine whether the
benefits of activation-informed perturbation geometry extend to
substantially larger autoregressive language models. In contrast to
RoBERTa-large, these experiments test PAMS in a setting where the
parameter dimension is considerably larger and the model architecture
and activation distributions differ from those of encoder-only
transformers.

% =========================================================
% OPT-2.7B RESULTS
% =========================================================
On OPT-2.7B, PAMS achieves the strongest overall performance among the
evaluated methods, with a mean accuracy of $71.94\%$, corresponding to
a $+2.07$ percentage-point gain over MeZO.
PAMS matches or outperforms other baselines  on five of the seven tasks, including
RTE, BoolQ, WiC, MultiRC, and COPA, with particularly notable gains
over MeZO on RTE ($+4.62$ points) and MultiRC ($+4.46$ points). Compared with LOZO and
HiZOO, PAMS improves the mean accuracy by $3.60$ and $5.87$ percentage
points, respectively. These results indicate that PAMS provides
consistent gains across diverse language understanding tasks while
maintaining the zeroth-order optimization setting.
% is competitive with MeZO, LOZO, and HiZOO and leads
% on RTE, BoolQ, WiC, and MultiRC. 
% Where it is not first, it is usually still close: 
% on SST-2, it trails LOZO by only $0.3$ points ($93.42$ vs.\ $93.73$),
% and on CB it is within $0.8$ of MeZO ($69.00$ vs.\ $69.77$). Its
% overall Mean ($68.16$) is likewise near MeZO's best mean
% ($68.84$). The main gaps are WSC and COPA, where MeZO remains clearly
% stronger; elsewhere PAMS either wins or is near the best reported
% score, supporting that activation-metric subspace ZO matches or nearly
% matches strong ZO baselines on most tasks.

On OPT-13B, PAMS achieves the highest mean performance among the
evaluated ZO methods on both task groups. On SST-2, RTE, CB, BoolQ,
and WSC, PAMS attains a mean accuracy of $72.54\%$, improving over
MeZO by $2.46$ percentage points and outperforming LOZO and HiZOO by
$2.65$ and $2.72$ points, respectively. PAMS improves over MeZO on
four of the five tasks, with particularly strong gains on RTE ($+2.47$
points) and BoolQ ($+2.57$ points).
On the remaining five tasks, PAMS achieves a mean of $74.84\%$, a
$0.57$-point gain over MeZO, and remains competitive with the
strongest baselines across individual tasks. Overall, these results
demonstrate that PAMS scales effectively to the 13B-parameter setting
and provides consistent improvements over existing zeroth-order
optimizers. 

% On OPT-2.7B, PAMS achieves the highest overall mean score ($68.96$) and leads on five of the eight tasks, although its aggregate advantage over MeZO ($68.84$) is modest. We view this result as complementary to, rather than inconsistent with, the substantially larger gains observed on RoBERTa-large and OPT-13B. PAMS does not assume that activation-informed anisotropy is equally informative in every model-task regime; rather, its benefit depends on how strongly the activation-derived covariance aligns with directions relevant to the downstream objective. The OPT-2.7B results suggest a regime in which isotropic perturbations are already competitive, leaving less headroom for additional geometric shaping. In contrast, PAMS improves the average RoBERTa-large score by more than two percentage points over the strongest baseline and exhibits substantially larger gains on OPT-13B, indicating that its advantage becomes pronounced when the underlying activation geometry provides a more informative preconditioning signal. This model- and task-dependent behavior is consistent with prior ZO benchmarking, which finds that optimizer effectiveness depends strongly on task alignment and on the trade-off between algorithmic complexity and optimization benefit ~\cite{pmlr-v235-zhang24ad}. Thus, the OPT-2.7B result delineates the regime in which the additional PAMS computation is less beneficial, while the RoBERTa-large and OPT-13B results demonstrate the regimes where activation-informed perturbations provide clear gains.

\subsection{Sensitivity to the PAMS Geometry}
\label{sec:exp_sensitivity}
\begin{figure*}[t]
    \centering
    % ---------- (a) Rank sensitivity ----------
    \begin{subfigure}[t]{0.31\textwidth}
        \centering
        \includegraphics[width=\linewidth]{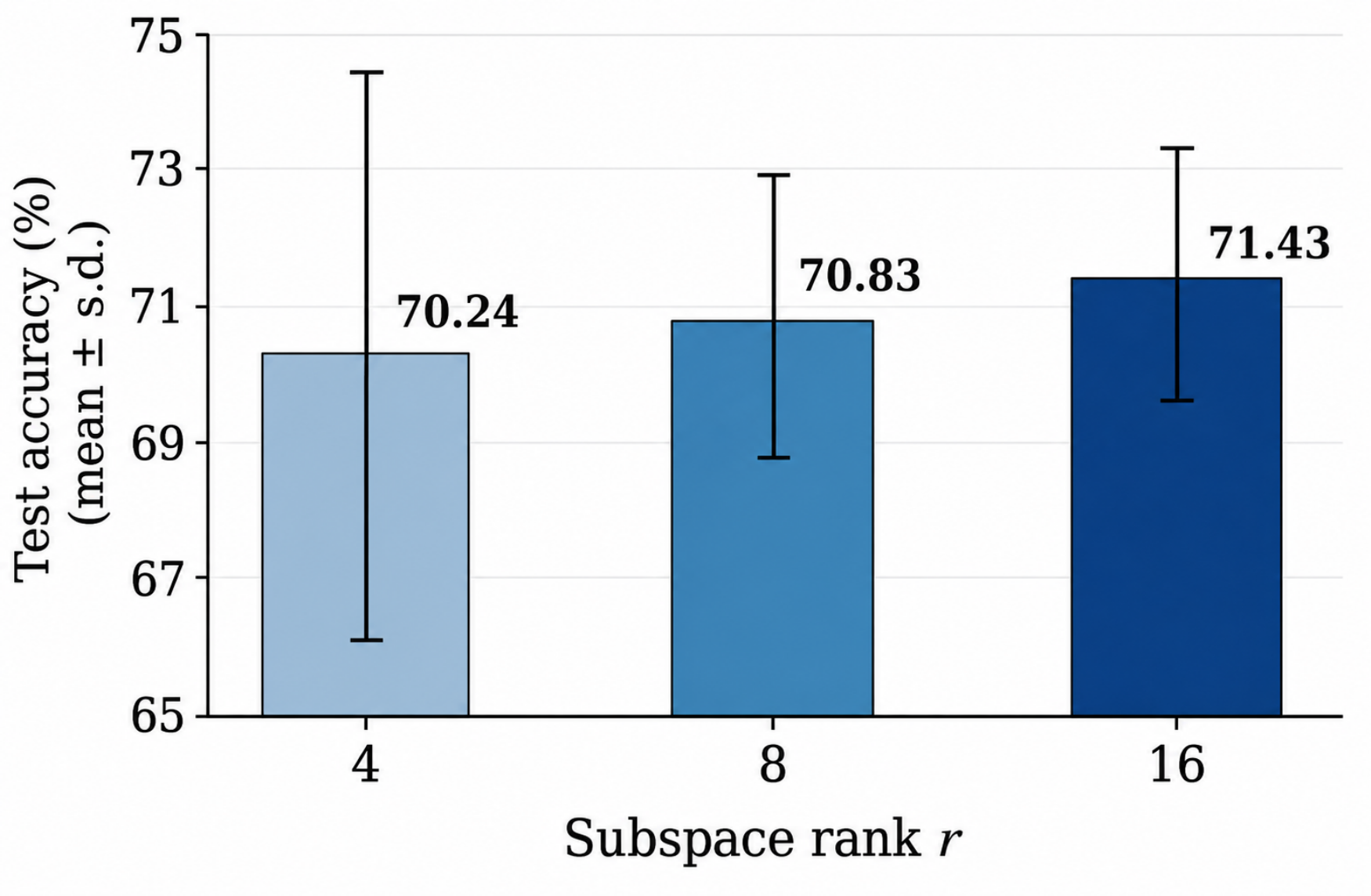}
        \caption{Sensitivity to activation-subspace rank $r$ on CB
        using OPT-13B.}
        \label{fig:rank_sensitivity}
    \end{subfigure}%
    \hfill
    % ---------- (b) Alpha sensitivity ----------
    \begin{subfigure}[t]{0.31\textwidth}
        \centering
        \includegraphics[width=\linewidth]{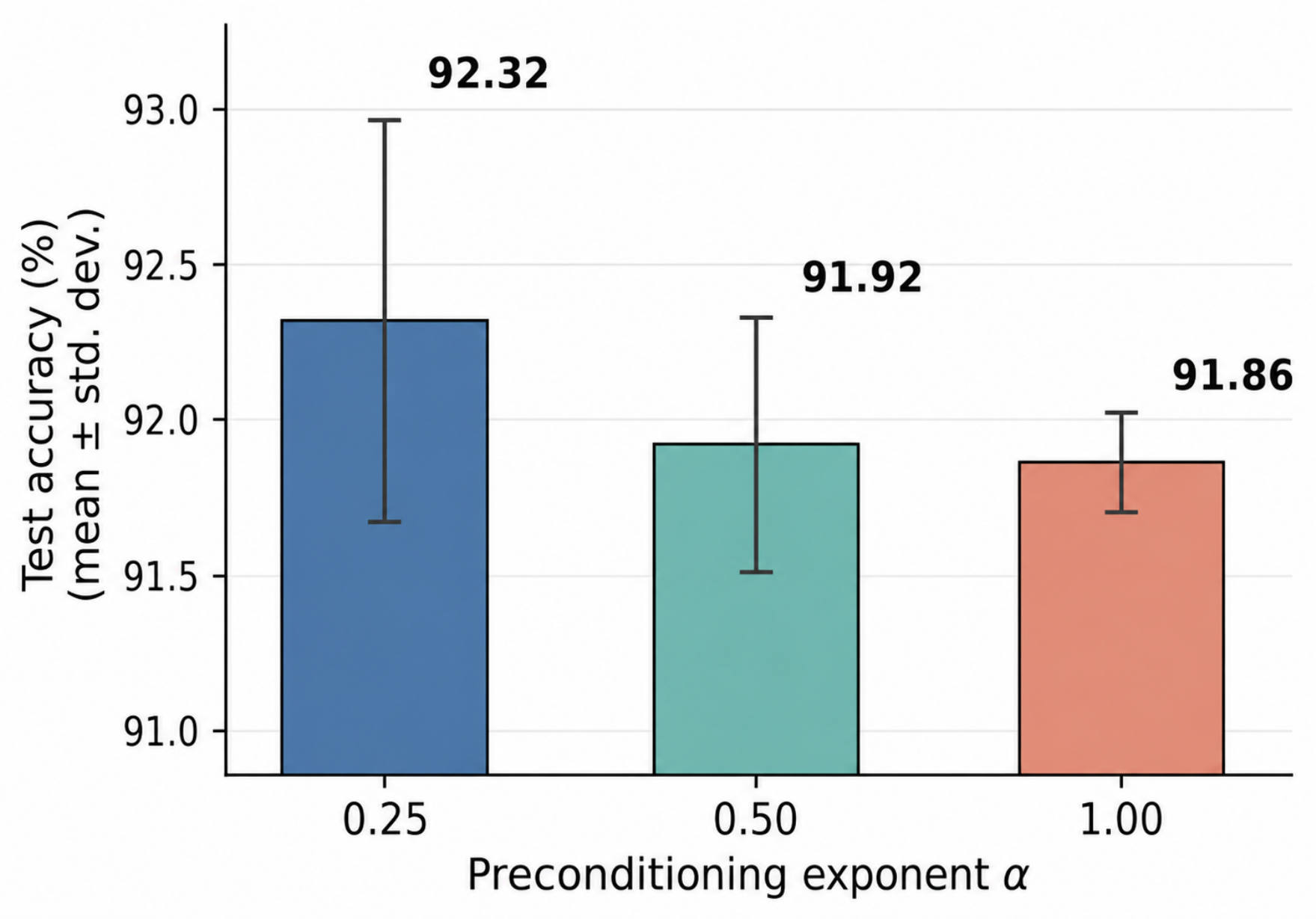}
        \caption{Sensitivity to preconditioning exponent $\alpha$ on
        SST-2 using RoBERTa-large.}
        \label{fig:alpha_sensitivity}
    \end{subfigure}%
    % ---------- (c) eta_U sensitivity (OPT-2.7B) ----------
    % Old (c) damping / rho panel:
    % \begin{subfigure}[t]{0.32\textwidth}
    %     \vspace{0pt}
    %     \centering
    %     \includegraphics[
    %         width=\linewidth,
    %         height=0.7\textwidth
    %     ]{figures/damping_factor_impact.png}
    %     \caption{Sensitivity to damping factor $\rho$ on SST-2 using
    %     RoBERTa-large.}
    %     \label{fig:rho_sensitivity}
    %   \end{subfigure}
    \hfill
    \begin{subfigure}[t]{0.31\textwidth}
      \centering
      \includegraphics[width=\textwidth]{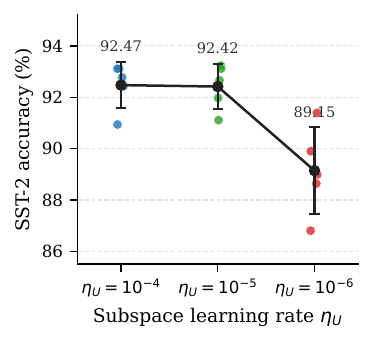}
      \caption{Sensitivity to subspace learning rate $\eta_U$ on SST-2
        using OPT-2.7B (BS$=$16).}
      \label{fig:eta_u_sensitivity}
    % ---------- (c) eta_U sensitivity ----------
    \end{subfigure}

    \caption{Sensitivity of PAMS to its principal perturbation-geometry
    hyperparameters: (a)~retained activation-subspace rank $r$,
    (b)~preconditioning exponent $\alpha$, and (c)~subspace learning rate
    $\eta_U$.
    In each experiment, all other PAMS hyperparameters are held fixed.
    For (a) and (b), bars report mean test accuracy across random seeds,
    with error bars denoting the sample standard deviation.
    For (c), points are per-seed accuracies and the line shows mean
    $\pm$ std on OPT-2.7B SST-2; accuracy is stable at
    $\eta_U\in\{10^{-4},10^{-5}\}$ and drops at $10^{-6}$.}
    \label{fig:pams_geometry_sensitivity}
\end{figure*}

% The PAMS perturbation distribution is principally controlled by the
% retained rank $r$, the preconditioning exponent $\alpha$, and the
% damping parameter $\rho$.
The retained rank $r$ determines the number of dominant
activation directions % s whose perturbation variances are modified,
subject to reweighting, while $\alpha$ controls its strength.
Figure~\ref{fig:rank_sensitivity} shows PAMS performance across
different values of $r$, with all other hyperparameters fixed. 
Higher ranks yield slightly better performance at the cost of
increased memory and computation.
Figure~\ref{fig:alpha_sensitivity} varies the preconditioning exponent
$\alpha$. Performance remains relatively stable across the tested
values, with $\alpha=0.25$ achieving the highest mean test accuracy,
suggesting that moderate preconditioning is preferable to aggressive
spectral attenuation. This observation is consistent with the broader
effectiveness of moderated spectral transformations in matrix-aware
optimizers such as \textsc{Shampoo}~\cite{Gupta2018ShampooPS} and
\textsc{Muon}~\cite{pmlr-v267-bernstein25a}.
Unlike these gradient-based methods, PAMS derives its geometry solely
from forward-pass activation statistics.
Finally, 
Figure~\ref{fig:eta_u_sensitivity} evaluates sensitivity to the
subspace learning rate $\eta_U$. Test accuracy remains nearly 
unchanged for $10^{-5}<\eta_U< 10^{-4}$, indicating that PAMS is
robust to the choice of $\eta_U$ within this range.

{\color{black}
\subsection{Ablation of the Activation Geometry}
\label{sec:exp_ablation}
% Table~\ref{tab:ablation} removes one part of the PAMS geometry at a
% time on RoBERTa-large.
Table~\ref{tab:ablation} investigates the contribution of individual
components of PAMS through ablation experiments on RoBERTa-large. Each
variant removes or freezes one component while the remaining settings
are kept unchanged.

\paragraph{Removing the reweighting ($\alpha=0$).}
Setting $\alpha=0$ yields $\mat{P}_l=\mat{I}$, 
resulting in isotropic perturbations. Removing the reweighting reduces
accuracy from 91.78\% to 89.78\% on SST-2 and from 67.29\% to
63.5\% on RTE. This performance degradation suggests that
activation-based reweighting helps capture the underlying activation
geometry, leading to more effective perturbations and improved
optimization performance compared with isotropic perturbations.

\paragraph{Freezing a random subspace ($\eta_U=0$).}
% With $\eta_U=0$ the initial subspace is held fixed, while the
% reweighting is kept at $\alpha=0.5$. On SST-2 at learning rate
% $10^{-3}$, this falls to chance, far below PAMS
% (Table~\ref{tab:ablation}).
Setting $\eta_U = 0$ freeze the initial random subspace while
retaining the reweighting factor $\alpha=0.5$. At a learning rate of
$10^{-3}$, accuracy on SST-2 drops to 49.2\%, compared with 91.78\%
for full PAMS. This substantial degradation indicates that maintaining
a fixed perturbation subspace is insufficient for effective training. 

\paragraph{Freezing the activation covariance.}
After a 100-step warmup, both the subspace and the spectrum of
$\mat{\Sigma}_A$ are held fixed, with $\alpha=0.5$ and learning rate
$10^{-3}$. On SST-5 this variant achieves only 17.1\% accuracy,
compared with 47.32\% for online PAMS.
% online PAMS (Table~\ref{tab:ablation}).

Overall, the ablation results highlight the importance of both
activation-based reweighting and online adaptation of the geometry.
Freezing the subspace or activation covariance substantially degrades
performance and can lead to training divergence without learning rate
adjustment.
% At the learning rate used for PAMS, both a frozen random subspace and a
% frozen activation covariance fail to train. The PAMS accuracies in
% Table~\ref{tab:roberta_results} therefore depend on a learned subspace
% that keeps tracking the activation covariance.

\begin{table}[t]
\centering
\caption{Ablation study of the activation geometry on RoBERTa-large. Test
accuracy (\%), reported as mean $\pm$ sample standard deviation over
five seeds. Each ablation removes or freezes one component of
activation geometry.
% The PAMS and MeZO columns are the results in
% Table~\ref{tab:roberta_results}.
}
\label{tab:ablation}
\resizebox{\columnwidth}{!}{%
% \begin{tabular}{llccc}
% \toprule
% Ablation & Task & Ablated & PAMS & MeZO \\
% \midrule
% $\alpha=0$ & SST-2 & $89.7\pm 2.1$ & $91.78\pm 1.09$ & $90.30\pm 1.0$ \\
% $\alpha=0$ & RTE & $63.5\pm 3.2$ & $67.29\pm 4.25$ & $62.46\pm 5.0$ \\
% $\eta_U=0$ & SST-2 & $49.2\pm 1.1$ & $91.78\pm 1.09$ & $90.30\pm 1.0$ \\
% Static $\mat{\Sigma}_A$ & SST-5 & $17.1\pm 6.2$ & $47.32\pm 0.82$ & $45.94\pm 2.5$ \\
% \bottomrule
% \end{tabular}
\begin{tabular}{llcc}
\toprule
Ablation Setting & Task & Ablated PAMS & Full PAMS \\
\midrule
No reweighting  & SST-2 & $89.7\pm 2.1$ & $91.78\pm 1.09$ \\
No reweighting  & RTE & $63.5\pm 3.2$ & $67.29\pm 4.25$ \\
Frozen Random subspace & SST-2 & $49.2\pm 1.1$ & $91.78\pm 1.09$  \\
Static Activation Covariance  & SST-5 & $17.1\pm 6.2$ & $47.32\pm 0.82$ \\
\bottomrule
\end{tabular}}
\end{table}
}
\subsection{Memory and Computational Efficiency}
\label{sec:exp_efficiency}
% An important objective of PAMS is to introduce structured
% perturbations without sacrificing the principal memory advantage of ZO
% fine-tuning.
% We therefore evaluate both accelerator-memory overhead
% and objective-evaluation cost. 
Table~\ref{tab:overhead_sst2} compares the per-iteration wall-clock
time and peak GPU memory usage on RoBERTa-large.
PAMS incurs $\mathcal{O}(n_l r)$ persistent memory and
$\mathcal{O}(n_l b r)$ additional computation per layer due to online
subspace tracking. Although PAMS has higher
per-iteration time than MeZO and LOZO, it remains faster than HiZOO.
The low-rank representation also keeps the memory overhead modest: PAMS
uses 2.16 GB of GPU memory, compared with 3.30 GB for HiZOO and 1.72
GB for MeZO.

\begin{table}[t]
\centering
\caption{Runtime, peak GPU memory, and test accuracy on RoBERTa-large
SST-2 ($K{=}16$, seed~42, ${\sim}$20k steps).}
\label{tab:overhead_sst2}
\begin{tabular}{lcccc}
\toprule 
 & MeZO & LOZO & HiZOO & PAMS \\
\midrule
% Wall-clock & 1.65\,h & 1.82\,h & 2.48\,h & 2.23\,h \\
Time / Step (ms) & 297 & 327 & 446 & 402 \\
Peak Mem. (GB) & 1.72 & 1.80 & 3.30 & 2.16 \\
Test acc & 91.05\% & 91.50\% & 91.05\% & 92.78\% \\
\bottomrule
\end{tabular}
\end{table}

\begin{figure}[t]
\centering
\includegraphics[width=\columnwidth]{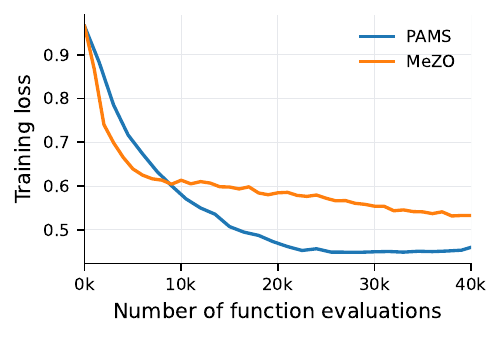}
\caption{Training loss on OPT-2.7B CB against function evaluations, at
learning rate $10^{-6}$. PAMS uses three function evaluations per
iteration and MeZO two; both are shown up to 40{,}000 evaluations.
Both curves start from a shared value, the mean loss over the first 50
iterations; later points are 500-iteration means. PAMS and MeZO uses
$\eta_U=10^{-6}$ on five seeds.}
\label{fig:opt27_cb_loss}
\end{figure}

Figure~\ref{fig:opt27_cb_loss} compares training performance of PAMS
and MeZO on the CB dataset using OPT-2.7B, under a matched computation
budget. Since PAMS performs three function evaluations
per iteration, compared with two for MeZO, we use the number of
function evaluations as the x-axis to ensure a fair comparison.
As shown in Figure~\ref{fig:opt27_cb_loss}, PAMS achieves lower
training loss with substantially fewer function evaluations. In
particular, PAMS reaches MeZO's final loss of 0.529 after only
15{,}000 function evaluations, using less than 40\% of the 40{,}000 required by MeZO.
With 40{,}000 evaluations, PAMS further reduces the loss to 0.460.
% at learning rate $10^{-6}$, against the number of function
% evaluations. A PAMS iteration uses three function evaluations and a
% MeZO iteration two, so both methods are shown up to MeZO's full budget
% of 40{,}000 evaluations. Both methods start from the same pretrained
% weights, so the first point is one shared value, the mean loss over the
% first 50 iterations; each later point is the mean over 500 iterations.
% PAMS falls below MeZO after about 9{,}000 evaluations and stays lower
% for the rest of the budget. It reaches MeZO's final loss of 0.529
% after 15{,}000 evaluations, less than 40\% of the 40{,}000 that MeZO
% needs, and ends the budget at 0.460.

% For OPT-13B, PAMS adds approximately 112.5~MiB of persistent FP32
% activation-subspace state. This state scales as
% $\mathcal{O}(n_lr)$ for a layer with input dimension $n_l$, rather
% than $\mathcal{O}(n_l^2)$ as would be required to explicitly maintain
% a dense activation second-moment matrix. Because no backward graph,
% parameter gradients, or optimizer moment tensors are required, the
% additional PAMS state remains small relative to the underlying model.

% \begin{table}[t]
% \centering
% \caption{Computational-efficiency comparison. Peak accelerator memory
% and wall-clock time should be measured under matched model, batch-size,
% and hardware settings.}
% \label{tab:efficiency}
% \small
% \begin{tabular}{lcccc}
% \toprule
% Method
% & Fwd./step
% & Bwd./step
% & Extra State
% & Peak Mem. \\
% \midrule
% MeZO
% & 2 & 0 & -- & -- \\

% HiZOO
% & -- & 0 & -- & -- \\

% LOZO
% & -- & 0 & -- & -- \\

% PAMS
% & 3 & 0 & 112.5 MiB$^\dagger$ & -- \\
% \bottomrule
% \end{tabular}

% \vspace{1mm}
% \footnotesize
% $^\dagger$Measured for the rank-16 PAMS state on OPT-13B.
% \end{table}

% In addition to peak memory, we report wall-clock training time and the number of objective evaluations required to reach a given
% internal-development performance whenever directly comparable runs
% are available. This distinction is necessary because different ZO
% optimizers may require different numbers of objective evaluations per
% parameter update.

%%% Local Variables:
%%% mode: latex
%%% TeX-master: "../main"
%%% End:

\section{Conclusions}
\label{sec:conclusion}
We presented Preconditioned Activation-Metric Subspace (PAMS), a
zeroth-order optimization framework for memory-efficient fine-tuning
of large language models. PAMS addresses a central limitation of
conventional ZO methods --- the use of isotropic perturbations --- by
exploiting geometric information that is directly available from
forward-pass activations. For each layer, PAMS tracks a low-rank
approximation of the dominant activation subspace and its associated
spectrum, and uses this information to construct a trace-normalized
perturbation covariance. The resulting perturbations softly reweight
exploration according to the activation spectrum while retaining
isotropic exploration outside the retained subspace. 

% Our analysis establishes that the perturbation covariance has a direct
% optimization interpretation: it induces right-side preconditioning of
% the gradient of the corresponding Gaussian-smoothed objective. This
% provides a principled connection between forward-pass activation
% statistics and zeroth-order optimization geometry, without requiring
% backpropagated gradients, Hessian information, or dense curvature
% matrices. The online subspace-tracking procedure further makes the
% approach practical for large transformer layers by maintaining only
% $\mathcal{O}(n_lr)$ state for a layer with feature dimension $n_l$ and
% retained rank $r$. 

% More broadly, PAMS demonstrates that gradient-free optimization need
% not rely on unstructured random exploration. Forward computation
% itself contains information that can be used to shape the search
% distribution, opening a broader direction for designing data- and
% model-aware zeroth-order optimizers. Future work could investigate
% adaptive rank and damping schedules, richer activation statistics
% beyond second moments, and extensions to other architectures and
% large-scale generative fine-tuning settings. 

%%% Local Variables:
%%% mode: latex
%%% TeX-master: "../main"
%%% End:

\section*{Acknowledgement}
This research was supported by the National Science Foundation under Grant No. 1943046.

% \section*{Acknowledgments} 
% This research was supported by the National Science Foundation under Grant No. 1943046.

% \balance % Keep normal column output before the appended complete proofs.
\bibliographystyle{template/IEEEtran}
\bibliography{IEEEabrv,main}

% \nobalance
% \input{subfiles/appendix}

\nobalance 
\clearpage
\appendices

\section{Deferred Proofs}
\label{app:proofs}
\begin{theoryaddition}
We use Assumptions~\ref{assump:lower_bounded}--\ref{assump:stochastic_oracle}
and the estimator in~\eqref{eq:minibatch_estimator}. Let
$\mathcal{F}_t$ denote the history before the fresh loss mini-batch,
$\mathbb{E}_t[\cdot]=\mathbb{E}[\cdot\mid\mathcal{F}_t]$, and
$\vec{g}_t=\grad\mathcal{L}(\vec{w}_t)$. The initial point is fixed. Write
$\Delta_0=\mathcal{L}(\vec{w}_0)-\mathcal{L}^{\star}$.
Throughout, $\tr(\mat{C}_t)=d$ and
$\kappa_P^{-1}\mat{I}\preceq\mat{C}_t\preceq\kappa_P\mat{I}$.
The covariance is $\mathcal{F}_t$-measurable in
Appendices~\ref{app:preconditioning}, \ref{app:bias}, and
\ref{app:convergence}. The second-moment argument also allows
$\mat{C}_t$ to depend on the current loss batch, as treated in
Appendix~\ref{app:dependent}; the Gaussian seed remains independent
after conditioning on that batch.
\end{theoryaddition}

\subsection{Proof of Lemma~\ref{lem:zo_precond}}
\label{app:preconditioning}
\begin{theoryaddition}
We fix $\vec{w}$ and $\mat{C}\succeq0$, and write
$\vec{u}=\mat{C}^{1/2}\vec{z}$ with
$\vec{z}\sim\Norm{\vec{0},\mat{I}}$. The smoothing is
\begin{equation}
 \mathcal{L}_{\varepsilon,\mat{C}}(\vec{x})
 =\E{\mathcal{L}(\vec{x}+\varepsilon\vec{u})}.
 \label{app:eq:smoothing}
\end{equation}
Every derivative of~\eqref{app:eq:smoothing} holds $\mat{C}$ fixed.
A globally $M$-Lipschitz gradient has at most linear growth, and
the corresponding function has at most quadratic growth. Hence all
Gaussian expectations below are finite, boundary terms vanish in
integration by parts, and differentiation may pass through the
smoothing expectation.
\end{theoryaddition}
Taking the expectation of the estimator gives
\begin{align}
  \E[\vec{u}]{\hvec{g}_{\varepsilon,\mat{C}}(\vec{w};\vec{u})}
  &=\frac{1}{2\varepsilon}\E[\vec{u}]{\left(\mathcal{L}(\vec{w}+\varepsilon\vec{u}) -
    \mathcal{L}(\vec{w}-\varepsilon\vec{u})\right)\vec{u}}\nonumber \\
  &=\frac{1}{\varepsilon}
    \E[\vec{u}]{\mathcal{L}(\vec{w}+\varepsilon\vec{u})\vec{u}}\,, \label{app:eq:antithetic_symmetry}
\end{align}
where the last equality uses the symmetry of the Gaussian
distribution, 
$\E[\vec{u}]{\mathcal{L}(\vec{w}-\varepsilon\vec{u})\vec{u}}=-\E[\vec{u}]{\mathcal{L}(\vec{w}+\varepsilon\vec{u})\vec{u}}$.
For $\vec{u}\sim\Norm{\vec{0}, \mat{C}}$, Stein\theoryadd{'s} identity states that, for
any sufficiently regular scalar function $f$, $\E[\vec{u}]{\vec{u}f(\vec{u})} =
\mat{C}\E[\vec{u}]{\grad_{\vec{u}}f(\vec{u})}$.
\begin{theoryaddition}
For a standard Gaussian coordinate, integration by parts gives
$\E{z_i h(\vec{z})}=\E{\partial_i h(\vec{z})}$, since
$z_i e^{-z_i^2/2}=-\frac{\mathrm{d}}{\mathrm{d}z_i}e^{-z_i^2/2}$.
Applying this coordinate-wise with $\vec{u}=\mat{C}^{1/2}\vec{z}$
and using the chain rule proves the covariance identity above.
\end{theoryaddition}
Applying this identity to $f(\vec{u}) =
\mathcal{L}(\vec{w}+\varepsilon\vec{u})$, we obtain
\begin{align*}
  \E[\vec{u}]{\hvec{g}_{\varepsilon,\mat{C}}(\vec{w};\vec{u})}
  &=\frac{\mat{C}}{\varepsilon}\E[\vec{u}]{\grad_{\vec{u}}\mathcal{L}(\vec{w}+\varepsilon\vec{u})}\\
  &=\frac{\mat{C}}{\varepsilon}\E[\vec{u}]{\varepsilon\grad_{\vec{w}}\mathcal{L}(\vec{w}+\varepsilon\vec{u})}   \\
  &=\mat{C}\grad_{\vec{w}}\E[\vec{u}]{\mathcal{L}(\vec{w}+\varepsilon\vec{u})}\,,
\end{align*}
where the penultimate equality follows from the chain rule and the
final equality follows by differentiating under
expectation. Therefore,
$\E[\vec{u}]{\hvec{g}_{\varepsilon,\mat{C}}(\vec{w};\vec{u})} =
\mat{C}\grad_{\vec{w}}\mathcal{L}_{\varepsilon, \mat{C}}(\vec{w})$.

\begin{theoryaddition}
For the stochastic estimator, condition on $\mathcal{F}_t$.
Predictability fixes both $\vec{w}_t$ and $\mat{C}_t$ before
$\xi_t$ is drawn. Independence of $\xi_t$ and $\vec{u}_t$ allows
each conditional mean loss to be replaced by $\mathcal{L}$, so
\begin{equation}
 \mathbb{E}_t[\hvec{g}_t]
   =\mat{C}_t\grad\mathcal{L}_{\varepsilon,\mat{C}_t}(\vec{w}_t).
 \label{app:eq:conditional_mean}
\end{equation}
The shared mini-batch includes all randomness of the stochastic loss.
The required interchange of expectations follows from sample
smoothness: $|\ell(\vec{x};\xi)|$ is bounded by
$|\ell(\vec{0};\xi)|+
\norm{\grad\ell(\vec{0};\xi)}\norm{\vec{x}}
+M\norm{\vec{x}}^2/2$. Loss integrability, the gradient second-moment
assumption, and finite Gaussian moments give absolute integrability
at the perturbed points.

For a layer block, column-wise vectorization satisfies
$\vect(\mat{G}\bmat{P})=(\bmat{P}^{\intercal}\otimes\mat{I})
\vect(\mat{G})$. Thus symmetry of $\bmat{P}_{l,t}$ gives
\begin{equation}
 \mathbb{E}_t[\hmat{G}_{l,t}]
  =\grad_{\mat{W}_l}\mathcal{L}_{\varepsilon,\mat{C}_t}
       (\mat{W}_t)\,\bmat{P}_{l,t}.
 \label{app:eq:layer_mean}
\end{equation}
The smoothing covariance in~\eqref{app:eq:layer_mean} is the
full-network covariance $\mat{C}_t$, not only its layer block.
No derivative is taken through a parameter-dependent covariance
selection rule.
\end{theoryaddition}

\subsection{Proof of Lemma~\ref{lem:smoothing_bias}}
\label{app:bias}
\begin{theoryaddition}
All expectations in this proof are over the Gaussian draw with
$\mat{C}_t$ held fixed.
\end{theoryaddition}

By $M$-smoothness,
\begin{align*}
  \norm{\grad{\mathcal{L}}_{\varepsilon,  \mat{C}_t}(\vec{w}_t) -
  \vec{g}_t}
  &=\norm{\E{\grad{\mathcal{L}(\vec{w}_t+\varepsilon\vec{u}_t)} -
    \grad{\mathcal{L}}(\vec{w}_t)}}\\
  &\leq M\varepsilon\E{\norm{\vec{u}_t}} \\
  &\leq M\varepsilon\sqrt{\tr(\mat{C}_t)}\,.
\end{align*}
Because of the trace normalization, we have
\[
  \tr(\mat{C}_t) = \sum_{l}m_l \tr(\bmat{P}_{l,t}) = \sum_{l}m_ln_l = d\,.
\]
Thus,
\[
  \norm{\grad{\mathcal{L}}_{\varepsilon,  \mat{C}_t}(\vec{w}_t) -
    \vec{g}_t}\leq M\varepsilon\sqrt{d}\,.
\]

\subsection{Proof of Lemma~\ref{lem:second_moment}}
\label{app:second_moment}
\begin{theoryaddition}
Conditioned on $(\mathcal{F}_t,\xi_t)$, the sample gradient
$\vec{g}_{t,\xi}=\grad\ell(\vec{w}_t;\xi_t)$ is fixed. Until the final
batch-averaging step, expectations in this proof are over the Gaussian
seed only. Both evaluations share $\xi_t$, and $\mat{C}_t$ is fixed
before drawing that seed, even when selected using $\xi_t$.
\end{theoryaddition}
Let $s_t = \frac{\theorydelmath{\mathcal{L}}\theoryadd{\ell}(\vec{w}_t+\varepsilon\vec{u}_t\theoryadd{;\xi_t}) -
  \theorydelmath{\mathcal{L}}\theoryadd{\ell}(\vec{w}_{\theorydelmath{-}\theoryadd{t}}\mathbin{\theorydelmath{+}\theoryadd{-}}\varepsilon\vec{u}_t\theoryadd{;\xi_t})}{2\varepsilon}$ and
$\hvec{g}_t = s_t \vec{u}_t$ denote the gradient estimate.
Let $\phi(z) = \theorydelmath{\mathcal{L}}\theoryadd{\ell}(\vec{w}_t + z\vec{u}_t\theoryadd{;\xi_t})$. Then $\phi'(z) =
\grad{\theorydelmath{\mathcal{L}}\theoryadd{\ell}}(\vec{w}_t + z\vec{u}_t\theoryadd{;\xi_t})^{\intercal}\vec{u}_t$. By
the fundamental theorem of calculus,
\begin{align*}
  \theorydelmath{\mathcal{L}}\theoryadd{\ell}(\vec{w}_t+\varepsilon\vec{u}_t\theoryadd{;\xi_t}) -
  \theorydelmath{\mathcal{L}}\theoryadd{\ell}(\vec{w}_t-\varepsilon\vec{u}_t\theoryadd{;\xi_t})
  &= \int_{-\varepsilon}^{\varepsilon}\phi'(z) \dd{z}
\end{align*}
Therefore,
\begin{align*}
  &\frac{\theorydelmath{\mathcal{L}}\theoryadd{\ell}(\vec{w}_t+\varepsilon\vec{u}_t\theoryadd{;\xi_t}) -
  \theorydelmath{\mathcal{L}}\theoryadd{\ell}(\vec{w}_t-\varepsilon\vec{u}_t\theoryadd{;\xi_t})}{2\varepsilon} -
  \vec{g}_{t\theoryadd{,\xi}}^{\intercal}\vec{u}_t\\
  &=\frac{1}{2\varepsilon}\int_{-\varepsilon}^{\varepsilon}\left[\grad{\theorydelmath{\mathcal{L}}\theoryadd{\ell}}(\theorydelmath{\vec{w}-t}\theoryadd{\vec{w}_t}
    + z\vec{u}_t\theoryadd{;\xi_t}) -
    \grad{\theorydelmath{\mathcal{L}}\theoryadd{\ell}}(\vec{w}_t\theoryadd{;\xi_t})\right]^{\intercal}\vec{u}_t \dd{z}\\
  &\abs*{\frac{\theorydelmath{\mathcal{L}}\theoryadd{\ell}(\vec{w}_t+\varepsilon\vec{u}_t\theoryadd{;\xi_t}) -
  \theorydelmath{\mathcal{L}}\theoryadd{\ell}(\vec{w}_t-\varepsilon\vec{u}_t\theoryadd{;\xi_t})}{2\varepsilon} -
    \vec{g}_{t\theoryadd{,\xi}}^{\intercal}\vec{u}_t}\\
  &\leq \frac{1}{2\varepsilon}\int_{-\varepsilon}^{\varepsilon}
    M\abs{z}\norm{\vec{u}_t}^{2}\dd{z} \\
  &=\frac{M\varepsilon}{2}\norm{\vec{u}_t}^{2}\,.
\end{align*}
Let $\gamma_t = s_t - \vec{g}_{t\theoryadd{,\xi}}^{\intercal}\vec{u}_t$. Then 
$\hvec{g}_t = (\vec{g}_{t\theoryadd{,\xi}}^{\intercal}\vec{u}_t + \gamma_t)\vec{u}_t$
with $\abs{\gamma_t}\leq \frac{M\varepsilon}{2}\norm{\vec{u}_t}^{2}$.
Taking the squared norm,
\begin{align*}
  \norm{\hvec{g}_t }^{2}
  &=(\vec{g}_{t\theoryadd{,\xi}}^{\intercal}\vec{u}_t
    +\gamma_t)^{2}\norm{\vec{u}_t}^{2}\\
  \intertext{Using the inequality $(a+b)^{2}\leq 2a^{2}+2b^{2}$,}
  \norm{\hvec{g}_t }^{2}
  &\leq  2(\vec{g}_{t\theoryadd{,\xi}}^{\intercal}\vec{u}_t)^{2} \norm{\vec{u}_t}^{2}+
    2\gamma_t^{2}\norm{\vec{u}_t}^{2}\,.
\end{align*}
We now bound each term separately. The first term can be bounded using
the fourth Gaussian moment.
For a Gaussian vector $\vec{u} \sim \Norm{\vec{0}, \mat{C}}$,
\begin{align*}
  \E{(\vec{g}^{\intercal}\vec{u})^{2}\norm{\vec{u}}^{\theoryadd{2}}}
  &=\E{\vec{g}^{\intercal}(\vec{u}\vec{u}^{\intercal})\vec{g} \cdot
    \tr(\vec{u}\vec{u}^{\intercal})} \\
  &=\E{\left(\sum_{i,j}g_ig_ju_iu_j\right) \left(\sum_k
    u_k^{2}\right)} \\
  &=\sum_{i,j,k}g_ig_j\E{u_iu_ju_k^{2}} \\
  \intertext{Using the Gaussian fourth-moment expansions for
  $\E{u_iu_ju_ku_k}$:}
  &=\sum_{i,j,k}g_ig_j(C_{ij}C_{kk} +2C_{ik}C_{jk})\\
  &=\sum_{i,j}g_ig_jC_{ij}\sum_{k}C_{kk} +
    \sum_{i,j,k}2g_ig_jC_{ik}C_{kj}\\
  &=(\vec{g}^{\intercal}\mat{C}\vec{g})\tr(\mat{C}) +
    2\vec{g}^{\intercal}\mat{C}^{2}\vec{g} 
\end{align*}
Thus, we get
\begin{align*}
  &\E{(\vec{g}_{t\theoryadd{,\xi}}^{\intercal}\vec{u}_t)^{2}\norm{\vec{u}_t}^{\theoryadd{2}}}\\
  &=(\vec{g}_{t\theoryadd{,\xi}}^{\intercal}\mat{C}_t\vec{g}_{t\theoryadd{,\xi}})\tr(\mat{C}_t)\\
  &\quad+2\vec{g}_{t\theoryadd{,\xi}}^{\intercal}\mat{C}_t^{2}\vec{g}_{t\theoryadd{,\xi}}\\
  &\leq \theorydelmath{c_{\max}}\theoryadd{\kappa_P}\norm{\vec{g}_{t\theoryadd{,\xi}}}^{2}\tr(\mat{C}_t) +
    2\theorydelmath{c_{\max}}\theoryadd{\kappa_P}^{2}\norm{\vec{g}_{t\theoryadd{,\xi}}}^{2}\\
  &=(d \theorydelmath{c_{\max}}\theoryadd{\kappa_P}+ 2\theorydelmath{c_{\max}}\theoryadd{\kappa_P}^{2})\norm{\vec{g}_{t\theoryadd{,\xi}}}^{2}
\end{align*}
\begin{theoryaddition}
For the sixth moment, we diagonalize $\mat{C}_t$ with eigenvalues
$c_i\geq0$. Then $\norm{\vec{u}_t}^2$ has the distribution of
$\sum_i c_i z_i^2$ for independent standard Gaussians. Gaussian
integration by parts gives $\E{z_i^{2q}}=(2q-1)\E{z_i^{2q-2}}$,
so the second, fourth, and sixth moments are $1$, $3$, and $15$.
Consequently,
\begin{align*}
 \E{\norm{\vec{u}_t}^6}
 &=15\sum_i c_i^3+9\sum_{i\ne j}c_i^2c_j
                   +6\sum_{i<j<k}c_ic_jc_k\\
 &=\tr(\mat{C}_t)^3+6\tr(\mat{C}_t)\tr(\mat{C}_t^2)
                   +8\tr(\mat{C}_t^3).
\end{align*}
Nonnegative eigenvalues imply
$\tr(\mat{C}_t^2)\leq\tr(\mat{C}_t)^2$ and
$\tr(\mat{C}_t^3)\leq\tr(\mat{C}_t)^3$.
\end{theoryaddition}
The second term is
\begin{align*}
  &2\gamma_t^{2}\norm{\vec{u}_t}^{2}\\
  &\leq 2 \frac{M^{2}\varepsilon^{2}}{4}\norm{\vec{u}_t}^{6} =
    \frac{M^{2}\varepsilon^{2}}{2}\norm{\vec{u}_t}^{6} \\
  \intertext{Taking the expectation,}
  &\E{2\gamma_t^{2}\norm{\vec{u}_t}^{2}}\\
  &\mathrel{\theorydelmath{=}\theoryadd{\leq}}\frac{M^{2}\varepsilon^{2}}{2}\E{\norm{\vec{u}_t}^{6}} \\
  &=\frac{M^{2}\varepsilon^{2}}{2}\left(\tr(\mat{C}_t)^{3}+6\tr(\mat{C}_t)\tr(\mat{C}_t^{2})
    + 8\tr(\mat{C}_t^{3})\right) \\
  &\leq \frac{15M^{2}\varepsilon^{2}}{2}\tr(\mat{C}_t)^{3} =
    \frac{15M^{2}\varepsilon^{2}d^{3}}{2}\,.
\end{align*}
Combining \theorydel{together}\theoryadd{these bounds}, we get
\[
  \E{\norm{\hvec{g}_t}^{2}} \leq 2\theorydelmath{c_{\max}}\theoryadd{\kappa_P}(d +
  2\theorydelmath{c_{\max}}\theoryadd{\kappa_P})\norm{\vec{g}_{t\theoryadd{,\xi}}}^{2} + \frac{15M^{2}\varepsilon^{2}d^{3}}{2}\,.
\]

\begin{theoryaddition}
Finally, we average over $\xi_t$. Unbiasedness and the variance
assumption gives
\begin{align*}
 \E[t]{\norm{\vec{g}_{t,\xi}}^2}
 &=\norm{\vec{g}_t}^2
   +2\vec{g}_t^{\intercal}\E[t]{\vec{g}_{t,\xi}-\vec{g}_t}\\
 &\quad+\E[t]{\norm{\vec{g}_{t,\xi}-\vec{g}_t}^2}\\
 &\leq\norm{\vec{g}_t}^2+\sigma_b^2.
\end{align*}
Thus
\begin{align}
 \E[t]{\norm{\hvec{g}_t}^2}
 &\leq2\kappa_P(d+2\kappa_P)(\norm{\vec{g}_t}^2+\sigma_b^2)
       \nonumber\\
 &\quad+\frac{15}{2}M^2d^3\varepsilon^2.
 \label{app:eq:minibatch_second_moment}
\end{align}
The Gaussian moment bounds hold conditionally on each $\xi_t$, so
this argument also covers covariance selected using the current batch.
\end{theoryaddition}

\subsection{Proof of Theorem~\ref{thm:convergence}}
\label{app:convergence}
\begin{theoryaddition}
Here $\mat{C}_t$ is $\mathcal{F}_t$-measurable, as required by
Theorem~\ref{thm:convergence}, and $\eta_t=\eta>0$ satisfies its
step-size restriction. Equation~\eqref{app:eq:conditional_mean}
therefore gives
$\E[t]{\hvec{g}_t}=\mat{C}_t(\vec{g}_t+\vec{b}_t)$.
All conditional expectations average over both the fresh batch and
the Gaussian seed.
\end{theoryaddition}

Since $\mathcal{L}$ is $M$-smooth,
\begin{align}
  \mathcal{L}(\vec{w}_{t+1})
  &\leq \mathcal{L}(\vec{w}_t) - \eta_t\ip{\vec{g}_t, \hvec{g}_t} +
    \frac{M\eta_t^{2}}{2}\norm{\hvec{g}_t}^{2} \nonumber \\
  \intertext{Taking conditional expectation,}
  \E[t]{\mathcal{L}(\vec{w}_{t+1})}
  &\leq \mathcal{L}(\vec{w}_t) - \eta_t\ip{\vec{g}_t, \E[t]{\hvec{g}_t}}
    + \frac{M\eta_t^{2}}{2}\E[t]{\norm{\hvec{g}_t}^{2}} \nonumber \\
  &= \mathcal{L}(\vec{w}_t) - \eta_t\ip{\vec{g}_t,
    \mat{C}_t\grad{\mathcal{L}}_{\varepsilon, \mat{C}_t}(\vec{w}_t)}
  \nonumber \\
  &\qquad +
    \frac{M\eta_t^{2}}{2}\E[\theoryadd{t}]{\norm{\hvec{g}_t}^{2}}\,. \label{app:eq:descent}
\end{align}
Let $\vec{b}_t = \grad{\mathcal{L}}_{\varepsilon,  \mat{C}_t}(\vec{w}_t) - \vec{g}_t$. Then,
\[
  \ip{\vec{g}_t, \mat{C}_t\grad{\mathcal{L}}_{\varepsilon,
      \mat{C}_t}(\vec{w}_t)}
  = \vec{g}_{\theoryadd{t}}^{\intercal}\mat{C}_t \vec{g}_t + \vec{g}_t^{\intercal}\mat{C}_t\vec{b}_t\,.
\]
By Assumption~\ref{assump:bounded_geometry},
$\vec{g}_t^{\intercal}\mat{C}_t\vec{g}_t\geq
\kappa_P^{-1}\norm{\vec{g}_t}^{2}$. Moreover,
$\abs{\vec{g}_t^{\intercal}\mat{C}_t\vec{b}_t} \leq
\kappa_P\norm{\vec{g}_t}\norm{\vec{b}_t}$. Using Young's inequality,
\[
  \kappa_P\norm{\vec{g}_t}\norm{\vec{b}_t}\leq
  \frac{1}{2\kappa_P}\norm{\vec{g}_t}^{2}+\frac{\kappa_P^{3}}{2}\norm{\vec{b}_t}^{2}\,.
\]
Thus,
\begin{equation}
  \label{app:eq:alignment}
    \ip{\vec{g}_t, \mat{C}_t\grad{\mathcal{L}}_{\varepsilon,
      \mat{C}_t}(\vec{w}_t)}
  \geq
  \frac{1}{2\kappa_P}\norm{\vec{g}_t}^{2}-\frac{\kappa_P^{3}M^{2}d\varepsilon^{2}}{2}\,.
\end{equation}
Substituting the gradient second-moment bound
in~\eqref{eq:second_moment_bound} and~\eqref{app:eq:alignment}
into~\eqref{app:eq:descent} yields
\begin{align*}
  &\E[t]{\mathcal{L}(\vec{w}_{t+1})}\\
  &\leq \mathcal{L}(\vec{w}_t) +
    \eta_t\left(\frac{\kappa_P^{3}M^{2}d\varepsilon^{2}}{2}-\frac{1}{2\kappa_P}\norm{\vec{g}_t}^{2}\right)\\  
  &\qquad
    +\frac{M\eta_t^{2}}{2}\left(2\kappa_P(d + 2\kappa_P)\theoryadd{\big(}\norm{\vec{g}_t}^{2}\theoryadd{+\sigma_b^2\big)}+\frac{15}{2}M^{2}d^{3}\varepsilon^{2}\right)\\
  &= \mathcal{L}(\vec{w}_t) 
    -\frac{\eta_t}{2}\left[\kappa_P^{-1} -
    2M\eta_t\kappa_P(d+2\kappa_P)\right]\norm{\vec{g}_t}^{2} \\
  &\qquad +
    \left[\frac{\kappa_P^{3}M^{2}d\varepsilon^{2}\eta_t}{2}
    +\frac{15M^3d^{3}\varepsilon^{2}\eta_t^{2}}{4}\right]\\
  &\qquad\theoryadd{+M\eta_t^2\kappa_P(d+2\kappa_P)\sigma_b^2}\,.
\end{align*}
If $\eta_t \leq \frac{1}{4M\kappa_P^{2}(d+2\kappa_P)}$, then $\kappa_P^{-1} -
2M\kappa_P(d+2\kappa_P) \eta_t\geq \frac{1}{2\kappa_P}$.
\begin{align*}
  &\E[t]{\mathcal{L}(\vec{w}_{t+1})}\\
  &\leq \mathcal{L}(\vec{w}_t) 
    -\frac{\eta_t}{4\kappa_P}\norm{\vec{g}_t}^{2} 
   +
    \left[\frac{\kappa_P^{3}M^{2}d\varepsilon^{2}\eta_t}{2}
    +\frac{15M^3d^{3}\varepsilon^{2}\eta_t^{2}}{4}\right]\\
  &\qquad\theoryadd{+M\eta_t^2\kappa_P(d+2\kappa_P)\sigma_b^2}\,.
\end{align*}
\begin{theoryaddition}
All total expectations used here are finite. Indeed,
$\norm{\grad\mathcal{L}(\vec{w})}^2
\leq2\norm{\grad\mathcal{L}(\vec{0})}^2+2M^2\norm{\vec{w}}^2$.
Together with~\eqref{eq:second_moment_bound} and
$\norm{\vec{w}_{t+1}}^2
\leq2\norm{\vec{w}_t}^2+2\eta^2\norm{\hvec{g}_t}^2$,
this proves finite second moments of the iterates by induction from
the fixed initial point. At-most-quadratic growth of $\mathcal{L}$
then ensures integrability of every objective value on a finite
horizon.
\end{theoryaddition}
Let $\eta_t = \eta$. Taking total expectations and summing from $t=0$ to $T-1$, we get
\begin{align*}
  \frac{\eta }{4\kappa_P}\sum_{t=0}^{T-1}\E{\norm{\vec{g}_t}^{2}}
  &\leq \mathcal{L}(\vec{w}_0) - \theorydelmath{\mathcal{L}_*}\theoryadd{\mathcal{L}^{\star}}\\
  &\qquad
     +
    TM^{2}d\varepsilon^{2}\eta\left[\frac{\kappa_P^{3}}{2} +
    \frac{15Md^{2}\eta}{4}\right]\\
  &\qquad\theoryadd{+TM\eta^2\kappa_P(d+2\kappa_P)\sigma_b^2}\,. \\
  \intertext{Dividing by $T\eta /(4\kappa_P)$ gives}
  \frac{1}{T}\sum_{t=0}^{T-1}\E{\norm{\vec{g}_t}^{2}}
  &\leq\frac{4\kappa_P(\mathcal{L}(\vec{w}_0) - \theorydelmath{\mathcal{L}_*}\theoryadd{\mathcal{L}^{\star}})}{\eta T}
    + 2M^{2}d \varepsilon^{2}\kappa_P^{\theorydelmath{3}\theoryadd{4}} \\
  &\qquad +
    15\kappa_PM^{3}d^{3}\varepsilon^{2}\eta\\
  &\qquad\theoryadd{+4M\eta\kappa_P^2(d+2\kappa_P)\sigma_b^2}\,.
\end{align*}

\begin{theoryaddition}
\subsection{Geometry Selected from the Current Loss Mini-Batch}
\label{app:dependent}
The same-batch analysis instead fixes $\vec{w}_t$ given
$\mathcal{F}_t$, draws a fresh $\xi_t$, selects
$\mat{C}_t=\mat{C}_t(\xi_t)$, and then draws an independent
standard Gaussian seed. The covariance satisfies the same almost-sure
bounds and is frozen during both loss evaluations. We prove
\eqref{eq:dependent_bound} with the quantities defined
in~\eqref{eq:dependence_residual}.
\begin{proof}
Conditioning on $(\mathcal{F}_t,\xi_t)$ fixes the covariance and
the sample loss. The integration-by-parts argument in
Appendix~\ref{app:preconditioning}, applied to that sample loss,
therefore gives
\begin{align}
 &\mathbb{E}[\hvec{g}_t\mid\mathcal{F}_t,\xi_t]\nonumber\\
 &\quad=\mat{C}_t\E[\vec{z}_t]{
   \grad\ell(\vec{w}_t+\varepsilon\mat{C}_t^{1/2}\vec{z}_t;\xi_t)}.
 \label{app:eq:sample_mean}
\end{align}
Let $\vec{b}_{t,\xi}$ denote the expectation on the right minus
$\vec{g}_{t,\xi}=\grad\ell(\vec{w}_t;\xi_t)$.
Sample smoothness and trace normalization give
$\norm{\vec{b}_{t,\xi}}\leq M\varepsilon\sqrt d$.
Consequently, with
$\vec{e}_t=\mathbb{E}_t[\mat{C}_t\vec{b}_{t,\xi}]$,
\begin{equation}
 \norm{\vec{e}_t}\leq\kappa_PM\varepsilon\sqrt d.
 \label{app:eq:sample_smoothing}
\end{equation}
Averaging~\eqref{app:eq:sample_mean} over $\xi_t$ and using
$\bmat{C}_t=\mathbb{E}_t[\mat{C}_t]$ yields
\begin{equation}
 \mathbb{E}_t[\hvec{g}_t]
   =\bmat{C}_t\vec{g}_t+\vec{r}_t+\vec{e}_t.
 \label{app:eq:dependent_mean}
\end{equation}
The second-moment proof in Appendix~\ref{app:second_moment}
already conditions on $\xi_t$ before applying the uniform Gaussian
moment bounds. It therefore proves the same
\eqref{eq:second_moment_bound} for this protocol.

Averaging preserves the spectral bounds, so
$\bmat{C}_t\succeq\kappa_P^{-1}\mat{I}$. Applying Young's
inequality separately to the two error terms gives
\begin{align*}
 \vec{g}_t^{\intercal}\mathbb{E}_t[\hvec{g}_t]
 &\geq\kappa_P^{-1}\norm{\vec{g}_t}^2
       -\norm{\vec{g}_t}(\norm{\vec{r}_t}+\norm{\vec{e}_t})\\
 &\geq\frac{\norm{\vec{g}_t}^2}{2\kappa_P}
          -\kappa_P\norm{\vec{r}_t}^2
          -\kappa_P\norm{\vec{e}_t}^2\\
 &\geq\frac{\norm{\vec{g}_t}^2}{2\kappa_P}
          -\kappa_P\norm{\vec{r}_t}^2
          -\kappa_P^3M^2d\varepsilon^2.
\end{align*}
We insert this bound into the conditional smoothness inequality
in~\eqref{app:eq:descent}, before its predictable-covariance
substitution, and apply the second-moment bound and the same
step-size restriction:
\begin{align}
 \mathbb{E}_t[\mathcal{L}(\vec{w}_{t+1})]
 &\leq\mathcal{L}(\vec{w}_t)
       -\frac{\eta}{4\kappa_P}\norm{\vec{g}_t}^2\nonumber\\
 &\quad+\eta\kappa_P\norm{\vec{r}_t}^2
       +\eta\kappa_P^3M^2d\varepsilon^2\nonumber\\
 &\quad+M\eta^2\kappa_P(d+2\kappa_P)\sigma_b^2\nonumber\\
 &\quad+\frac{15}{4}M^3d^3\eta^2\varepsilon^2.
 \label{app:eq:dependent_descent}
\end{align}
The finite-horizon moment argument above applies unchanged.
Let $R_T=T^{-1}\sum_{t=0}^{T-1}\E{\norm{\vec{r}_t}^2}$.
Taking total expectations, telescoping, and dividing by
$T\eta/(4\kappa_P)$ now gives
\begin{align}
 \frac1T\sum_{t=0}^{T-1}\E{\norm{\vec{g}_t}^2}
 &\leq\frac{4\kappa_P\Delta_0}{\eta T}
       +4M^2d\varepsilon^2\kappa_P^4\nonumber\\
 &\quad+4\kappa_P^2R_T
       +4M\eta\kappa_P^2(d+2\kappa_P)\sigma_b^2\nonumber\\
 &\quad+15M^3d^3\varepsilon^2\eta\kappa_P,
 \label{app:eq:dependent_stationarity}
\end{align}
which is~\eqref{eq:dependent_bound}.

Finally, unbiasedness implies
\[
 \vec{r}_t=\mathbb{E}_t\big[
  (\mat{C}_t-\bmat{C}_t)(\vec{g}_{t,\xi}-\vec{g}_t)\big].
\]
The operator-norm inequality and conditional Cauchy--Schwarz yield
\begin{align}
 \norm{\vec{r}_t}
 &\leq\mathbb{E}_t\big[
       \norm{\mat{C}_t-\bmat{C}_t}_{\mathrm{op}}
       \norm{\vec{g}_{t,\xi}-\vec{g}_t}\big]\nonumber\\
 &\leq\sigma_b\sqrt{
       \mathbb{E}_t\norm{\mat{C}_t-\bmat{C}_t}_{\mathrm{op}}^2}.
 \label{app:eq:residual_control}
\end{align}
This also ensures finiteness of $R_T$ under the uniform spectral
bounds. The dependence residual need not vanish as
$\varepsilon\to0$. For this upper bound to establish a vanishing
average squared gradient norm, the residual term must also vanish.
\end{proof}
\end{theoryaddition}

%%% Local Variables:
%%% mode: latex
%%% TeX-master: "../main"
%%% End:

\end{document}